\documentclass[sigconf]{acmart}
\AtBeginDocument{%
  }

\setcopyright{acmlicensed}
\copyrightyear{2018}
\acmYear{2018}
\acmDOI{XXXXXXX.XXXXXXX}
\acmConference[Conference acronym 'XX]{Make sure to enter the correct
  conference title from your rights confirmation email}{June 03--05,
  2018}{Woodstock, NY}
\acmISBN{978-1-4503-XXXX-X/2018/06}

\usepackage{subcaption}
\usepackage{multirow}
\usepackage{booktabs}
\usepackage{amsmath}
\usepackage{amsfonts}
\usepackage{enumitem}
\usepackage[normalem]{ulem}
\useunder{\uline}{\ul}{}
\usepackage{graphicx}
\usepackage{microtype}
\usepackage{verbatim}
\usepackage{lipsum}
\usepackage{algorithm}
\usepackage{algorithmic}
\usepackage{amsthm}
\theoremstyle{plain}
\newtheorem{theorem}{Theorem}

\newtheorem{definition}{Definition}

\begin{document}

\title{Hypergraph Contrastive Learning for both Homophilic and Heterophilic Hypergraphs}

\author{Renchu Guan}
\authornote{Key Laboratory of Symbolic Computation and Knowledge Engineering of the Ministry of Education}
\affiliation{%
  \institution{College of Computer Science and Technology, Jilin University}
  \city{Changchun}
  \country{China}
}
\email{guanrenchu@jlu.edu.cn}

\author{Xuyang Li}
\authornotemark[1]
\affiliation{%
  \institution{College of Computer Science and Technology, Jilin University}
  \city{Changchun}
  \country{China}
}
\email{xuyang24@mails.jlu.edu.cn}

\author{Yachao Zhang}
\authornotemark[1]
\affiliation{%
  \institution{College of Computer Science and Technology, Jilin University}
  \city{Changchun}
  \country{China}
}
\email{zyc25@mails.jlu.edu.cn}

\author{Wei Pang}
\affiliation{%
  \institution{School of Mathematical and Computer Sciences, Heriot-Watt University}
  \city{Edinburgh}
  \country{UK}
}
\email{w.pang@hw.ac.uk}

\author{Fausto Giunchiglia}
\affiliation{%
  \institution{Department of Information Engineering and Computer Science, University of Trento}
  \city{Trento}
  \country{Italy}
}
\email{fausto.giunchiglia@unitn.it}

\author{Ximing Li}
\authornotemark[1]
\affiliation{%
  \institution{College of Computer Science and Technology, Jilin University}
  \city{Changchun}
  \country{China}
}
\email{liximing86@gmail.com}

\author{Yonghao Liu}
\authornotemark[1]
\authornote{Corresponding Author}
\affiliation{
  \institution{College of Computer Science and Technology, Jilin University}
  \city{Changchun}
  \country{China}
}
\email{yonghao20@mails.jlu.edu.cn}

\author{Xiaoyue Feng}
\authornotemark[1]
\authornotemark[2]
\affiliation{%
  \institution{College of Computer Science and Technology, Jilin University}
  \city{Changchun}
  \country{China}
}
\email{fengxy@jlu.edu.cn}

\renewcommand{\shortauthors}{Renchu Guan et al.}
\begin{abstract}

Hypergraphs, as a generalization of traditional graphs, naturally capture high-order relationships. In recent years, hypergraph neural networks (HNNs) have been widely used to capture complex high-order relationships. However, most existing hypergraph neural network methods inherently rely on the homophily assumption, which often does not hold in real-world scenarios that exhibit significant heterophilic structures. To address this limitation, we propose \textbf{HONOR}, a novel unsupervised \textbf{H}ypergraph c\textbf{ON}trastive learning framework suitable for both hom\textbf{O}philic and hete\textbf{R}ophilic hypergraphs. Specifically, HONOR explicitly models the heterophilic relationships between hyperedges and nodes through two complementary mechanisms: a prompt-based hyperedge feature construction strategy that maintains global semantic consistency while suppressing local noise, and an adaptive attention aggregation module that dynamically captures the diverse local contributions of nodes to hyperedges. Combined with high-pass filtering, these designs enable HONOR to fully exploit heterophilic connection patterns, yielding more discriminative and robust node and hyperedge representations. Theoretically, we demonstrate the superior generalization ability and robustness of HONOR. Empirically, extensive experiments further validate that HONOR consistently outperforms state-of-the-art baselines under both homophilic and heterophilic datasets. 

\end{abstract}



\maketitle

\section{Introduction}

Hypergraph is the generalized graph structure, where hyperedges can connect any number of vertices, breaking the limitation of traditional graphs where edges can only connect two vertices. This enables hypergraphs to flexibly model multi-modal relationships and higher-order interactions \cite{lee2024survey, wu2024collaborative, lee2023m}, making them widely applicable in domains such as recommender systems \cite{xia2022self, xia2021self}, computer vision \cite{kim2022equivariant, jiang2022hypergraph} and bioinformatics \cite{zheng2019gene, wang2021multi}. In recent years, hypergraph neural networks (HNNs) have attracted widespread attention due to their powerful representation capabilities \cite{antelmi2023survey, li2024hypergraph}. 
Modern HNNs typically follow a two-stage message-passing paradigm \cite{antelmi2023survey, kim2024survey}, where information is first aggregated from nodes to hyperedges and then propagated back from hyperedges to nodes. Notably, the effectiveness of this mechanism heavily relies on the homophily assumption---that nodes connected by the same hyperedge share similar labels or semantic representations \cite{wang2022equivariant, li2025hypergraph}. This assumption, however, does not always hold in real-world scenarios. For example, in hypergraphs constructed from social media discussions, a single topic (\textit{i.e.}, hyperedge) may bring together users with opposing viewpoints, such as both environmental advocates and critics. When conventional HNNs are directly applied to such heterophilic hypergraphs, they tend to produce suboptimal node representations, ultimately degrading downstream task performance.

\begin{figure}[h]
    \begin{flushleft}
        \includegraphics[width=0.45\textwidth]{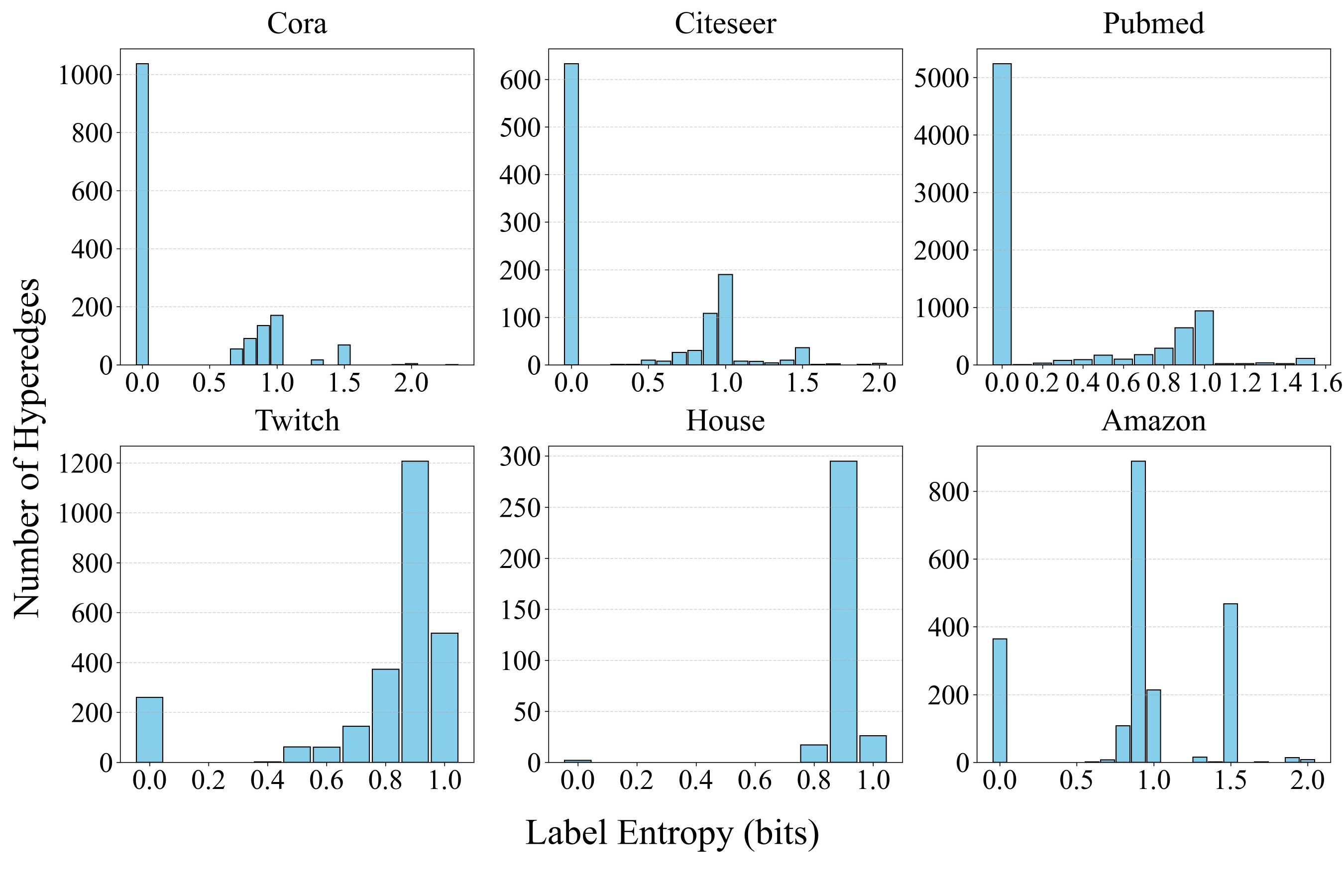}
    \end{flushleft}
    \caption{Statistics of label entropy on real-world datasets.}
    \label{intro}
\end{figure}

To precisely quantify the heterophily of real-world hypergraphs, we propose a novel heterophily discrimination metric called \textit{Label Entropy}, denoted as $H_{\mathcal{H}}$, 
which measures the diversity of node labels within each hyperedge. Its detailed definition can be found in Section \ref{definition}. 
This metric reflects the degree of heterophily in the label distribution within each hyperedge. A higher entropy value indicates greater diversity in node labels and stronger heterophily. We employ this metric to evaluate the heterophily of six real-world hypergraph datasets, as illustrated in Fig. \ref{intro}. It is evident that several datasets (\textit{e.g.}, Twitch, House, and Amazon) exhibit pronounced heterophily, indicating that the conventional homophily assumption may be insufficient to capture the complex and diverse structures present in such hypergraphs.

Currently, heterophilic graph learning has been extensively studied, giving rise to a variety of representation learning methods tailored for heterophilic structures \cite{chien2020adaptive, wang2022acmp, he2023contrastive, xiao2022decoupled}. However, heterophilic hypergraphs pose greater challenges in modelling high-order relationships, especially in unsupervised scenarios, where effective research and methodological frameworks are still lacking.
Hypergraph contrastive learning (HGCL) is emerging as an unsupervised learning paradigm \cite{kim2024survey}, enhancing representational capabilities through contrastive learning mechanisms. However, existing HGCL approaches also rely on the homophily assumption \cite{li2025hypergraph} and primarily utilize low-pass filtering operations to smooth neighbor node representations. While this is effective for homophilic structures, it inevitably leads to oversmoothing in heterophilic connections, thus obscuring important differences among nodes. An intuitive approach is to replace the low-pass filter with a high-pass filter. However, it is important to note that although using high-pass filtering alone can amplify specificity, it may introduce noise and weaken the stability of the global structure. Therefore, balancing multi-frequency domain information, effectively aggregating node and hyperedge information is one of the key challenges in heterophilic HGCL.


To this end, we propose a novel unsupervised HGCL framework, namely \textbf{HONOR}, which overcomes the limitations of relying on the homophily assumption, and can effectively handle both homophilic and heterophilic hypergraphs. The core idea is to explicitly model the heterophilic relationships between hyperedges and their connected nodes, enabling hyperedges and nodes to have discriminative representational capacity in feature space. Specifically, we construct contrastive views by leveraging two complementary hyperedge feature construction mechanisms that effectively capture the heterophilic connection patterns between nodes and hyperedges. First, to maintain global consistency within each hyperedge while suppressing noise and semantic variance, we design a prompt-based hyperedge feature construction strategy. Concretely, we compute a central pattern feature for each hyperedge and adjust the original node features by incorporating the cosine similarity between each node and the corresponding central feature. The adjusted node features are then used to generate prompt information, which is subsequently fused with the initial hyperedge features to produce prompt-enhanced hyperedge representations. This approach effectively captures the overall semantics of each hyperedge while mitigating the influence of local noise. Second, to capture the diverse local contributions of nodes to hyperedges, we introduce an adaptive attention aggregation mechanism that assigns dynamic weights to nodes, enabling the model to learn their true influence on hyperedges. Based on this mechanism, we construct hyperedge representations that complement the prompt-based features. The above designs successfully construct heterophilic membership relationships and enable the effective propagation of prompt-based and dynamic hyperedge features solely through high-frequency filters. This allows our model to fully exploit the heterophilic connection patterns between nodes and hyperedges in hypergraphs, thereby enhancing the representational distinctiveness. 
Moreover, we have demonstrated the effectiveness of the proposed model not only through empirical evaluations but also by theoretically proving its superior generalization and robustness. In summary, our main contributions are as follows.

\noindent $\bullet$ We propose the first unsupervised HGCL framework named HONOR tailored for heterophilic hypergraphs, addressing the limitations of existing approaches that predominantly focus on homophilic structures. 

\noindent $\bullet$ We demonstrate that HONOR exhibits strong adaptability, serving as a unified and generalizable representation learning framework for both homophilic and heterophilic hypergraphs. 

\noindent $\bullet$ We introduce a novel heterophily discrimination metric named Label Entropy to quantitatively assess the degree of heterophily in hypergraph structures.

\noindent $\bullet$ We theoretically unveil the underlying working mechanism of the proposed model and demonstrate that it exhibits superior generalization and robustness compared to existing approaches.

\noindent $\bullet$ Extensive experiments on multiple homophilic and heterophilic hypergraph datasets show HONOR consistently outperforms existing mainstream methods. 

\section{Related Work}

\noindent \textbf{Hypergraph Neural Network.}
HNNs learn node or hyperedge representations based on hypergraph structures to capture high-order information. This capability has recently attracted increasing attention. For example, HGNN \cite{feng2019hypergraph} is the first to apply neural network structures to hypergraphs, capturing high-order information by fusing nodes within hyperedges to enhance node representation capabilities. HNHN \cite{dong2020hnhn} further improves this by alternately updating node and hyperedge representations through interactive message passing, enabling richer high-order relational modeling. UniGNN \cite{huang2021unignn} proposes a unified framework that generalizes the message-passing mechanisms of various hypergraph neural networks into a consistent form, enabling flexible and efficient representation learning for hypergraph structures. However, these methods rely on supervised learning for model training.

\noindent \textbf{Hypergraph Contrastive Learning.}
To eliminate the dependency on label information, recent studies \cite{wei2022augmentations, lee2023m, li2024hypergraph} have explored combining contrastive learning with HNNs. For example, HyperGCL \cite{wei2022augmentations} designs a variational autoencoder to generate views for unsupervised high-quality representation learning. TriCL \cite{lee2023m} introduces tri-directional contrastive learning to simultaneously capture node-, group-, and membership-level information within hypergraphs. 
However, these hypergraph models are all designed under the homophily assumption.
In heterophilic graphs or hypergraphs, simply aggregating neighbor features may instead introduce noise and even degrade model performance.

\noindent \textbf{Heterophilic Graph Neural Network.}
In heterophilic graphs, the nodes connected by each edge often have different features or labels. To address the challenges of representation learning in heterophilic graphs, a series of supervised models have emerged in recent years \cite{abu2019mixhop, zhu2020beyond}. For example, 
H2GCN \cite{zhu2020beyond} combines high-order neighborhood modeling with feature retention mechanisms. 
Additionally, to handle real-world scenarios where labels are scarce, several contrastive learning methods for heterophilic graphs have been proposed \cite{liu2023beyond, yuan2023muse}. For instance, 
MUSE \cite{yuan2023muse} designs semantic and contextual views and employs a fusion controller to integrate multi-view information. 
However, compared with these methods for heterophilic graphs, learning on heterophilic hypergraphs have not been explored in depth and requires further research.
\section{Preliminary Study}
\label{definition}
\textbf{Notations.} For a hypergraph $\mathcal{H}=(\mathcal{V},\mathcal{E})$, $\mathcal{V}=\{v_1,v_2,\ldots,v_N\}$ denotes the set of nodes, and $\mathcal{E}=\{e_1,e_2,\ldots,e_M\}$ denotes the set of hyperedges. Each hyperedge is a non-empty subset of $\mathcal{V}$. The hypergraph $\mathcal{H}$ can be represented by an incidence matrix \( \mathbf{H} \in \mathbb{R}^{N \times M} \), where the matrix element \( \mathbf{H}_{ij} = 1 \) if node \( v_i \) belongs to hyperedge \( e_j \), and \( \mathbf{H}_{ij} = 0 \) otherwise.

To better model the heterophilic connectivity patterns between nodes and hyperedges,
we transform the hypergraph $\mathcal{H}$ into a bipartite graph $\mathcal{B}=(\mathcal{V} \cup \mathcal{E}, \mathcal{M})$, where $\mathcal{M}$ denotes the membership relations: each membership edge links a node to one of its incident hyperedges. The corresponding bipartite adjacency matrix is constructed as $\mathbf{A}=\left[\begin{array}{cc}\mathbf{0} & \mathbf{H} \\
\mathbf{H}^{\top} & \mathbf{0}
\end{array}\right] \in \mathbb{R}^{(N+M) \times(N+M)}, \text { where } \mathbf{H} \in\{0,1\}^{N \times M}$. The entry \( {\mathbf{A}}_{ij} \) takes the value 1 if there exists a membership relation between node \( v_i \) and hyperedge \( e_j \), and \( {\mathbf{A}}_{ij} = 0 \) otherwise. The symmetric normalized adjacency matrix is denoted as $\hat{\mathbf{A}}=\mathbf{D}^{-1/2}\mathbf{A}\mathbf{D}^{-1/2}$, where the degree matrix $\mathbf{D}$ is defined as $d(v_i)=\sum_{j=1}^{N+M}{\mathbf{A}}_{ij}$. The symmetric normalized Laplacian matrix is $\hat{\mathbf{L}}=\mathbf{I}-\hat{\mathbf{A}}$, where $\mathbf{I}$ is the identity matrix. Additionally, $\mathbf{X}\in\mathbb{R}^{(N+M)\times r}$  is the feature matrix of the nodes and the hyperedges, where $r$ is the feature dimension.\\
\textbf{Hypergraph Heterophily Metrics.} Considering that there are currently few metrics available to measure hypergraph heterophily, to quantify hypergraph heterophily, we employ two metrics, \textit{label entropy} proposed in this work, and \textit{heterophilic pairwise ratio} adapted from previous work \cite{peigeom, li2025hypergraph}. Their formal definitions are provided below.
\begin{definition}[Label Entropy]
    It is computed by calculating the entropy of the label distribution for each hyperedge, then normalized by the maximum possible entropy given the number of unique labels in that hyperedge, finally, the normalized label entropy values of all hyperedges are averaged, defined as follows:
    \begin{equation}
    H_{\mathcal{H}}=\frac{1}{|\mathcal{E}|}\sum_{e\in \mathcal{E}}\left(\frac{H(e)}{\log|C_e|}\right)=\frac{1}{|\mathcal{E}|}\sum_{e\in E}\left(\frac{-\sum_{c\in C_e}p(c)\log p(c)}{\log|C_e|}\right),
    \end{equation}
    where $C_e$ denotes the set of label classes within a hyperedge $e$, and $p(c)$ represents the occurrence probability of category $c$ within that hyperedge. A higher entropy value indicates a more diverse label distribution, implying greater heterophily in the hypergraph. 
\end{definition}

\begin{definition}[Heterophilic Pairwise Ratio]
    It measures the proportion of node pairs within hyperedges that belong to different label categories, reflecting the heterophily of node labels inside each hyperedge, defined as follows: 
    \begin{equation}
    R_{\mathcal{H}}=\frac{1}{|\mathcal{E}|}\sum_{e\in E}\left(\frac{1}{|e|(|e|-1)}\sum_{u<v}\mathbb{I}[y_u\neq y_v]\right),
    \end{equation}
    where $\mathbb{I}[\cdot]$ denotes the indicator function, which equals 1 when the node-pair has different labels, and 0 otherwise. A higher value indicates a greater number of cross-class node pairs within hyperedges, and thus stronger overall heterophily of the hypergraph.
\end{definition}
\section{Method}

\begin{figure*}[t]
    \centering    \includegraphics[width=0.95\textwidth]{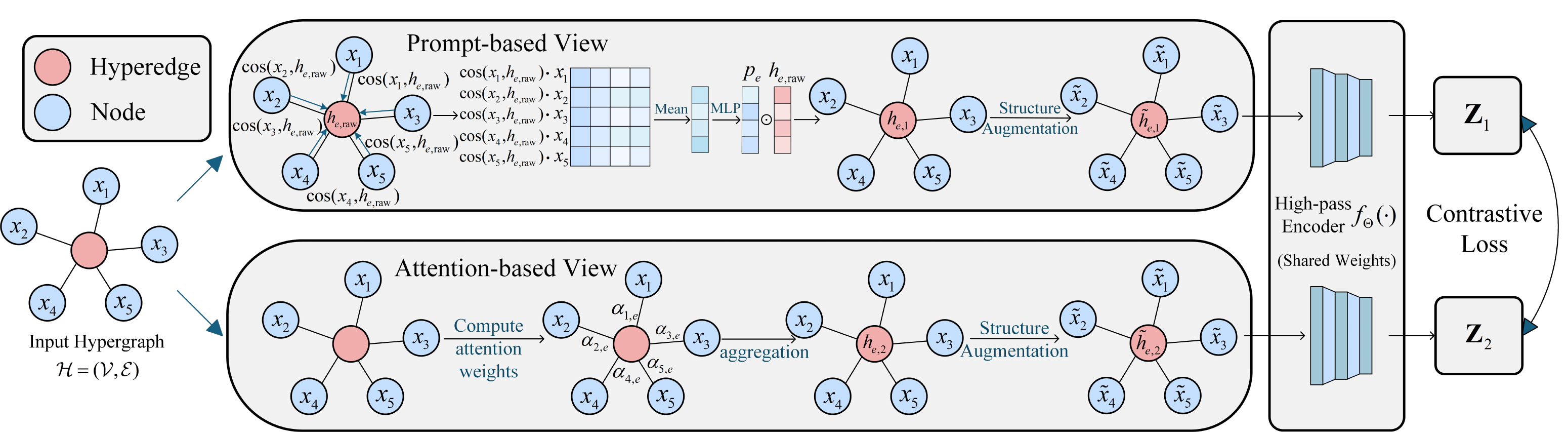}
    \caption{The overall framework of \textbf{HONOR}.}
    \label{Framework}
\end{figure*}

In this section, we detail the proposed HONOR model. The overall framework consists of three modules: \textit{hypergraph augmentation}, \textit{hypergraph encoder}, and \textit{model training}, as shown in Fig. \ref{Framework}.

\subsection{Hypergraph Augmentation}
To address the issues of complex and diverse node-hyperedge connection patterns and potential noise interference in heterophilic hypergraphs, we propose a view generation scheme based on prompt regulation and adaptive attention regulation, which effectively augments the raw hypergraph and provides complementary views for subsequent encoding and contrastive learning.

\subsubsection{Prompt-based View Generation.}

In heterophilic hypergraphs, each hyperedge often connects multiple nodes, which may exhibit significant semantic differences. This to some extent leads to non-homophily of information within hyperedges. Simply aggregating all node features equally to construct a hyperedge representation may introduce noise and irrelevant information. Thus, how to selectively augment context-consistent node information while suppressing interfering signals within each hyperedge is key to effective hyperedge feature modeling.

Inspired by the idea of prompt learning \cite{liu2023pre, su2023graph}, we introduce a prompt-based mechanism that leverages local similarity to enhance 
the hyperedge representations. Unlike traditional HNNs that apply a uniform aggregation for all nodes or hyperedges, this mechanism allows each hyperedge to generate a unique prompt vector based on its internal heterophily, thereby capturing high-order heterophilic connection patterns.
Specifically, for any hyperedge $e$, we first compute a raw local feature by averaging the features of its connected nodes, as follows:
\begin{equation}
h_{e,\mathrm{raw}}=\frac{1}{|e|}\sum_{v_i\in e}x_i.
\end{equation}
Then, we calculate the cosine similarity between each node and the hyperedge to reweight the node features. The final prompt factor $p_e$ is derived by feeding the weighted sum into a lightweight MLP for non-linear transformation:
\begin{equation}
p_e = \mathrm{MLP} \Bigg( 
    \frac{1}{|e|} \sum_{v_i \in e} 
    \cos\big(x_i,\, h_{e,\mathrm{raw}}\big) \cdot x_i 
\Bigg).
\end{equation}
Finally, in View 1, an element-wise product combines the prompt vectors with the raw hyperedge features to produce the prompt-based hyperedge representation:
\begin{equation}
\label{he1}
h_{e,1} = h_{e,\mathrm{raw}} \odot p_e.
\end{equation}
This mechanism effectively acts as a local consistency filter, which can identify context-consistent signals while suppressing the interference of inconsistent nodes on hyperedge features.

\subsubsection{Adaptive Attention-based View Generation.}

In heterophilic hypergraphs, there are often complex diversity dependencies between nodes and hyperedges, and a single similarity screening cannot cover this potential high-order heterophily. Therefore, we introduce an adaptive attention mechanism that learns node-specific contributions to hyperedges in a learnable manner, complementing the prompt-based view. Concretely, for any hyperedge $e$, we employ a trainable scoring function $\varphi_\theta(\cdot)$ to assign an importance score to each node feature, which is then normalized by a softmax to produce the final attention weight:
\begin{equation}
\alpha_{i,e} = 
\frac{
\exp\!\big( \varphi_\theta(x_i) \big)
}{
\sum_{v_j \in e} \exp\!\big( \varphi_\theta(x_j) \big)
}.
\end{equation}
In View 2, the final hyperedge representation under adaptive attention is then computed by the weighted aggregation:
\begin{equation}
\label{he2}
h_{e,2}=\sum_{v_i\in e}\alpha_{i,e}x_i.
\end{equation}
Unlike prompt-based mechanisms that depend on heuristic similarity measures, our adaptive attention regulation dynamically allocates information weights according to the intrinsic connection patterns within each hyperedge. This approach prevents the oversight of heterophilic node contributions that typically occurs with uniform aggregation schemes.

Furthermore, 
we introduce a simple feature dropout operation to prevent overfitting. Concretely, we randomly drop dimensions of the node and hyperedge features with a given rate $p$. Meanwhile, this approach naturally generates two similar views for each ordinal node. It encourages the encoder to learn representations that remain invariant to partial feature perturbations.

\subsubsection{Structural Feature Augmentation.}

In addition to feature-based augmentation methods, we introduce a simple yet effective structural augmentation strategy to further enrich the local structural information of nodes and hyperedges. Specifically, node degree is one of the most fundamental structural attributes and has been widely shown to provide important local connectivity information without introducing additional learnable parameters. Numerous studies \cite{zhu2023heterophily,ma2021homophily,yuan2023muse} have demonstrated the critical role of node degree in capturing structural information.

We assume that the degree of a node is the number of hyperedges connected to that node, and the degree of a hyperedge is the number of nodes connected by that hyperedge. To ensure comparability across different scales, we normalize the degree as follows:
\begin{equation}
\tilde{d}_i=\frac{d_i-d_{\min}}{d_{\max}-d_{\min}},
\end{equation}
where $d_{\min}$ and $d_{\max}$ are the minimum and maximum degrees observed in the current batch or the entire hypergraph. The normalized degree is then concatenated as an additional scalar feature:
\begin{equation}
\label{degree}
\tilde{x}_i=[x_i;\tilde{d}_i], \quad \tilde{h}_{e,1}=[h_{e,1};\tilde{d}_i], \quad\tilde{h}_{e,2}=[h_{e,2};\tilde{d}_i].
\end{equation}
Incorporating node and hyperedge degree provides explicit cues about local connection scales, helping HONOR capture structural importance. This design augments representation under heterophilic hypergraphs, especially under sparse structural conditions.
\subsection{Hypergraph Encoder}
In the hypergraph encoding stage, our objective is to effectively capture the high-order structural information embedded within the heterophilic connectivity patterns formed between nodes and hyperedges. Unlike traditional HNNs that typically adopt a two-stage message propagation paradigm, 
we explicitly model the given hypergraph as a bipartite graph consisting of ordinary nodes and hyperedge nodes. This bipartite formulation enables unified message propagation over all nodes, which explicitly include both ordinary nodes and hyperedge nodes, within a single graph structure.

In contrast to low-pass convolution operations, we introduce high-pass filtering into the encoder $f_\Theta(\cdot)$ to better preserve the heterophily and discriminative signals inherent in membership relations. Specifically, for the \( \ell \)-th layer, the propagation rule for the two views $\mathcal{H}_1$ and $\mathcal{H}_2$ is formulated as:
\begin{equation}
\label{high-pass}
\mathbf{X}_k^{(\ell)} = \sigma \Big[ \big(\mathbf{I} - \gamma^{(\ell)} \hat{\mathbf{L}} \big) \cdot \mathbf{W}^{(\ell)} \mathbf{X}_k^{(\ell-1)} \Big],
\end{equation}
where $\gamma^{(\ell)}$ denotes the predefined high-pass coefficient at layer $\ell$, $\mathbf{W}^{(\ell)}$ and $\sigma(\cdot)$ denote the trainable weights and the activation function, respectively. Moreover, \( \mathbf{X}_k^{(\ell)} \) represent the feature embeddings of view $k$ at layer $\ell$. When $\ell=0$, $\mathbf{X}_k^{(0)}=\left[\{\tilde{x}_i\}_{i=1}^N; \{\tilde{h}_{e,k} | e\in\mathcal{E}\}\right] \in\mathbb R^{(N+M)\times (r+1)}$. After encoding the hypergraph, we can derive the encoded embeddings $\mathbf{Z}_k$ of each view $k$.

The two high-pass filtering operations enable both nodes and hyperedge nodes to propagate high-frequency local difference signals within the unified bipartite structure, thereby enhancing the discriminative ability of the learned embeddings under heterophilic connectivity patterns.

\subsection{Model Training}
We design multiple objective functions to simultaneously enhance view alignment and suppress representation redundancy and noise interference. 
The overall loss function consists of the following three components:

\subsubsection{Contrastive Loss.}

For each entity \(i\) (which can be either a node or a hyperedge), we treat its embedding in the first view $z_{1}(i)$ as the anchor, its corresponding embedding in the second view $z_{2}(i)$ as the positive sample, and all other embeddings in the second view $z_{2}(j)$ (\(j \neq i\)) as negative samples. The similarity between embeddings is measured using cosine similarity, defined as $s(p,q)=\frac{p^\top q}{\|p\|\|q\|}$. The loss function for each positive sample pair is defined as follows:
\begin{equation}
\mathcal{L}_c(z_1(i),z_2(i))=-\log\frac{\exp\left(s(z_1(i),z_2(i))/\tau\right)}{\sum_{j=1}^{N+M}\exp\left(s(z_1(i),z_2(j))/\tau\right)},
\end{equation}
where $\tau$ is a temperature parameter. The objective function is defined as the average over positive pairs, given as follows:
\begin{equation}
\label{constrast}
\mathcal{L}_{\mathrm{contrast}}=\frac{1}{2(N+M)}\sum_{i=1}^{N+M}\left[\mathcal{L}_c(z_1(i),z_2(i))+\mathcal{L}_c(z_2(i),z_1(i))\right].
\end{equation}
\subsubsection{Structural Decoupling Loss.}

Considering that hypergraphs may contain redundant or noisy connections, we introduce a structural decoupling constraint based on hyperedge–node similarity. The core idea is to minimize unnecessary similarity between node and hyperedge representations so that the model focuses on true heterophilic relations. The loss can be formulated as follows:
\begin{equation}
\label{decoupling}
\mathcal{L}_{\mathrm{decouple}}=\sum_{(v,e)\in\mathcal{M}}\left(s(z_v,z_e)-t_{ve}\right)^2,
\end{equation}
where $z_v$ and $z_e$ denote the representations of node $v$ and hyperedge $e$, respectively, and $t_{ve}$ is the target similarity. In our setting, we simply set $t_{ve} = 0$.

\subsubsection{Covariance Regularization.}

To prevent the learned representations from degenerating into redundant or trivial low-rank patterns during training, we introduce a covariance regularization from a frequency-domain perspective, which constrains the spectral energy distribution. Specifically, we perform a Discrete Fourier transform (DFT) on both node and hyperedge embeddings and minimize the total spectral power. This encourages greater diversity in node and hyperedge features and enhances the expressiveness of heterophilic information. The regularization term is defined as follows:
\begin{equation}
\label{covariance}
\mathcal{L}_{\mathrm{cov}}=\|\mathrm{FFT}(\mathbf{Z}_v)\|_F^2+\|\mathrm{FFT}(\mathbf{Z}_e)\|_F^2,
\end{equation}
where $\mathbf{Z}_v$ and $\mathbf{Z}_e$ denote the embedding matrices of all nodes and hyperedges, respectively, $\left\|\cdot\right\|_{F}$ denotes the Frobenius norm, and $\mathrm{FFT}(\cdot)$ indicates the one-dimensional discrete Fourier transform along the feature dimension.

The final training objective combines the contrastive loss, structural decoupling loss, and covariance regularization:
\begin{equation}
\label{over_loss}
\mathcal{L}=\mathcal{L}_{\mathrm{contrast}}+\lambda_{1}\mathcal{L}_{\mathrm{decouple}}+\lambda_{2}\mathcal{L}_{\mathrm{cov}},
\end{equation}
where $\lambda_{1}$ and $\lambda_{2}$ denote the weight for $\mathcal{L}_{\mathrm{decouple}}$ and $\mathcal{L}_{\mathrm{cov}}$, respectively.

After model training, the final representation for downstream tasks is obtained by linearly combining the high-pass filtered embeddings from the two views, denoted as $\mathbf{Z}_1$ and $\mathbf{Z}_2$:
\begin{equation}
\label{opt}
\mathbf{Z} = \mathbf{Z}_{1} + \lambda_{3} \mathbf{Z}_{2},
\end{equation}
where $\lambda_{3}$ is a coefficient controlling the contribution of each view. 

This fused representation preserves the heterophilic structural information captured by prompt-based and adaptive attention-based views, which can be beneficial for various downstream tasks, such as node classification and node clustering. We present the training procedure in 
\textbf{Appendix} A.1 to facilitate a deeper understanding of HONOR. Moreover, the complexity analysis of HONOR are provided in \textbf{Appendix} A.2.

\section{Theoretical Analysis}


In this section, we provide a theoretical analysis to explain why HONOR is effective. 

To simplify the proof process, we consider a two-community hypergraph generated by the Hypergraph Stochastic Block Model (HSBM) \cite{kim2018stochastic} as the analytical setting. 
Let the model output be
\begin{equation}
\mathbf{Z} = \Delta(\Sigma_{\text{prompt}} + \lambda_3 \Sigma_{\text{attn}}) \mathbf{X},
\end{equation}
where $\Delta = (\mathbf{I} - \gamma^{(1)} \mathbf{L})(\mathbf{I} - \gamma^{(2)} \mathbf{L})$, $\mathbf{L}$ is the normalized Laplacian, $\gamma^{(1)}, \gamma^{(2)} > 0$, $\Sigma_{\text{prompt}}$ denotes the feature-aggregation matrix corresponding to the prompt-based view, and $\Sigma_{\text{attn}}$ denotes the feature-aggregation matrix corresponding to the attention-based view.
\begin{theorem}
\label{theorem1}
For a community \( C_k \) (\(k = 1, 2\)), we denote \( \mathbf{{Z}_{C_k}} \) as the submatrix of $\,$\( \mathbf{Z} \) consisting of the embeddings of all nodes in \( C_k \). The community mean embedding is defined as $\mu_{C_k} = \frac{1}{|C_k|} \sum_{i \in C_k} z_i$. We define the projections of $\,$$\mathbf{Z}$ onto the direction of the second smallest eigenvector $u_2$ and its orthogonal complement as $\mathrm{Proj}_{u_2} \mathbf{Z}$ and $\mathrm{Proj}_{u_2}^\perp \mathbf{Z}$, respectively. Then for class separation, the following inequality holds:
\begin{equation}
\|\mathbf{{Z}_{C_1}} - \mathbf{{Z}_{C_2}}\|_F \geq \Omega\left( (\frac{\alpha-\beta}{\beta}) \sqrt{N} \right),
\end{equation}
where $\left\|\cdot\right\|_{F}$ is the class separation, i.e., the Frobenius norm, $\Omega(\cdot)$ denotes the asymptotic lower bound, $N$ denotes the number of nodes, $\alpha$ represents the connection probability of hyperedges that connect only nodes within the same community, while $\beta$ represents the connection probability of hyperedges that span nodes from different communities. For the Signal-to-Noise Ratio (SNR) lower bound, we have the following:
\begin{equation}
\text{SNR} = \frac{\| \text{Proj}_{u_2} \mathbf{Z} \|_2}{\| \text{Proj}_{u_2}^{\perp} \mathbf{Z} \|_2} \geq \Omega\left( \frac{\alpha - \beta}{\beta} \right).
\end{equation}
\end{theorem}
Theorem \ref{theorem1} indicates that regardless of whether the hypergraph is homophilic or heterophilic, as long as the gap $\alpha - \beta$ is maintained, the model can ensure that the community embeddings are significantly separable in terms of both their means and overall distribution. This yields a sufficiently high SNR, allowing different communities to be clearly distinguishable in the embedding space.

\begin{theorem}
\label{theorem2}
When the modified Laplacian $\hat{\mathbf{L}}$ with prompt and attention weights is only a small perturbation of the original Laplacian $\mathbf{L}$, the gap between the second smallest eigenvector $\hat{u}_2$ and the true eigenvector $u_2$ is strictly bounded, we have the following perturbation error bound holds:
\begin{equation}
\|\hat{u}_2-u_2\|_2\leq\frac{\min(\phi_{\min},\alpha_{\min})}{(\alpha-\beta)^2}\cdot O\left(\sqrt{\frac{\log N}{N}}\right),
\end{equation}
where $\phi_{\min}$ denotes the minimum lower bound of the hyperedge weight $\phi_e$ and $\alpha_{\min}$ denotes the minimum lower bound of the node--hyperedge association weight $\alpha_{i,e}$.
\end{theorem}
Theorem \ref{theorem2} indicates that even with the introduction of variable weights from prompt and attention mechanisms, as long as these weights do not degenerate, the second smallest eigenvector, which is critical in the spectral embedding, can still be stably recovered with negligible perturbation. 

\begin{theorem}
\label{theorem3}
Let the two views after high-pass filtering be denoted as $\mathbf{Z}_1$ and $\mathbf{Z}_2$. Y represents the community labels, where $Y$ is binary. Then the following inequality holds:
\begin{equation}
I\left(Y ; \mathbf{Z}_1, \mathbf{Z}_2\right)>\max \left\{I\left(Y ; \mathbf{Z}_1\right), I\left(Y ; \mathbf{Z}_2\right)\right\},
\end{equation}
\begin{equation}
\mathcal{E}_{\mathrm{gen}} \leq O\left(\sqrt{\frac{\log N}{N}}\right)+O\left(N^{-1 / 2}\right)=O\left(\sqrt{\frac{\log N}{N}}\right),
\end{equation}
where \( I(\cdot\,; \cdot) \) is the mutual information and $\mathcal{E}_{\mathrm{gen}}$ denotes the generalization error of a single classifier on new samples.
\end{theorem}
Theorem \ref{theorem3} indicates that $\mathbf{Z}$ considers both the prompt-based view and the attention-based view, therefore it can contain more task-relevant information compared to considering either one alone. Simultaneously, the generalization error decreases with the increase in sample size $N$ and is inversely related to the mutual information shared by the two views.
The detailed proofs of Theorems \ref{theorem1}-\ref{theorem3} can be found in \textbf{Appendix} A.3.
\section{Experiments}

\noindent \textbf{Datasets.}
To evaluate the performance of HONOR, we use four homophilic and four heterophilic benchmark  hypergraph datasets:
\textit{Homophilic hypergraph datasets}:
(1) \textit{co-citation datasets} (\textbf{Cora-C}, \textbf{Citeseer}, and \textbf{PubMed}) \cite{sen2008collective}, and
(2) \textit{animals dataset} (\textbf{Zoo} \cite{asuncion2007uci});
\textit{Heterophilic hypergraph datasets}:
(1) \textit{political membership dataset} (\textbf{House} \cite{chien2021you}),
(2) \textit{movie-actor dataset} (\textbf{IMDB} \cite{wang2019heterogeneous}), 
(3) \textit{co-create dataset} (\textbf{Twitch} \cite{li2025hypergraph}), and
(4) \textit{co-purchasing dataset} (\textbf{Amazon} \cite{li2025hypergraph}). Detailed statistics, including Label Entropy $H_{\mathcal{H}}$ and Heterophilic Pairwise Ratio $R_{\mathcal{H}}$, are presented in Table~\ref{tab:t1}. In \textbf{Appendix} A.4, we give detailed descriptions of datasets.

\noindent \textbf{Baselines.}
We compare HONOR with representative \textit{supervised} and \textit{unsupervised} baselines. For the supervised models, we choose two
types: \textit{graph supervised learning} models and \textit{heterophilic graph supervised learning} models. The former contains \textbf{GCN} \cite{kipf2016semi}, \textbf{GAT} \cite{velickovic2017graph}, \textbf{HGNN} \cite{feng2019hypergraph}, \textbf{HyperGCN} \cite{yadati2019hypergcn}, \textbf{HNHN} \cite{dong2020hnhn}, \textbf{UniGCN} \cite{huang2021unignn}, and \textbf{AllSet} \cite{chien2021you}, while the latter includes \textbf{FAGCN} \cite{bo2021beyond}, \textbf{H2GCN} \cite{zhu2020beyond}, and \textbf{$R^{2}$LP} \cite{cheng2024resurrecting}. For the unsupervised models, we choose three types: \textit{graph contrastive learning} models, \textit{heterophilic graph contrastive learning} models, and \textit{hypergraph contrastive learning} models. The first category includes \textbf{DGI} \cite{velivckovic2018deep}, \textbf{GRACE} \cite{zhu2020deep}, \textbf{BGRL} \cite{thakoor2021bootstrapped}, and \textbf{COSTA} \cite{zhang2022costa}; the second category includes \textbf{GREET} \cite{liu2023beyond} and \textbf{MUSE} \cite{yuan2023muse}; while the third category includes \textbf{HyperGRL} \cite{du2022self}, \textbf{HyperGCL} \cite{wei2022augmentations}, \textbf{TriCL} \cite{lee2023m}, \textbf{HypeBoy} \cite{kim2024hypeboy}, and \textbf{MMACL} \cite{lee2024multi}. We present the descriptions of these baselines in \textbf{Appendix} A.5.

\begin{table}[h]
\centering
\caption{Statistics of the evaluated datasets.}
\label{tab:t1} 
\resizebox{0.46\textwidth}{!}{%
\begin{tabular}{@{}c|cccc|cccc@{}}
\toprule
\multirow{2}{*}{Dataset} & \multicolumn{4}{c|}{homophilic hypergraph datasets}
& \multicolumn{4}{c}{heterophilic hypergraph datasets}  \\ \cmidrule{2-9}
& Cora-C & Citeseer & PubMed & Zoo & House  & IMDB  & Twitch & Amazon \\
\midrule
\#Nodes    & 1,434 & 1,458     & 3,840 & 101  & 1,290  & 3,939  & 16,356  & 3,124 \\
\#Hyperedges           & 1,579 & 1,079     & 7,963  & 43  & 341  & 2,015  & 2,627 & 2,090 \\
\#Memberships          & 4,786 & 3,453     & 34,629   & 1,717  & 11,843   & 9,560  & 16,356 & 6,486 \\
\#Classes              & 7    & 6        & 3     & 7 & 2   & 3  & 2 & 5   \\
\#Features             & 1,433 & 3,703     & 500 & 16  & 2 & 3,066    & 7 & 111    \\
$H_{\mathcal{H}}$ & 0.3139 & 0.3795 & 0.2888 & 0.5890 & 0.9704 & 0.6471 & 0.8227 & 0.7756 \\
$R_{\mathcal{H}}$ & 0.2538 & 0.3186 & 0.2235 & 0.4692 & 0.5149 & 0.5121 & 0.5143 & 0.6323 \\
\bottomrule
\end{tabular}
}
\end{table}

\noindent \textbf{Evaluation Protocol.}
To evaluate the quality of the representations learned by HONOR, we select two widely used downstream tasks: node classification and node clustering. For the node classification task, we first train the high-pass hypergraph encoder in an unsupervised manner and freeze its parameters. Then, we use the trained model to randomly split the node representations into training, validation, and test sets with proportions of 10\%, 10\%, and 80\%, respectively, following the methodology of prior works \cite{lee2023m, wei2022augmentations}. Finally, we train a logistic regression classifier using the learned node embeddings and evaluate the performance on the test set. To ensure the reliability of the experiments, we perform 20 random splits for each dataset and report the average results along with their standard deviations. For the node clustering task, we apply the k-means clustering algorithm to the output node representations to obtain the clustering results. To evaluate clustering performance, we used common metrics such as Normalized Mutual Information (NMI) and Adjusted Rand Index (ARI), where higher values indicate better performance. We conduct k-means clustering five times on the generated node embeddings and report the average results.

\noindent \textbf{Implementation Details.}
We set the number of high-pass filtering $f_\Theta(\cdot)$ layers to 2 ($\ell$ in Eq.\ref{high-pass}). In all experiments, we employ Glorot initialization \cite{glorot2010understanding} and Adam optimizer \cite{kingma2014adam} for model training. All important hyperparameter settings, such as the temperature $\tau$ and the weight of structure decoupling loss $\lambda_{1}$, can be found in \textbf{Appendix} A.6. For classical graph neural network models such as GCN and GAT, as well as graph contrastive learning models like DGI and GRACE, they cannot be directly applied to hypergraphs. Therefore, following previous work settings \cite{lee2023m}, we convert the hypergraph into a clique-expanded graph for representation learning. For the remaining baselines, we implement them using the optimal hyperparameters reported by the authors in their original papers. All the experiments are conducted on the 24GB NVIDIA GeForce 3090Ti GPU with the Python 3.11 and PyTorch 2.0.1 environment.
\section{Results}

\begin{table*}[ht]
\tiny
\centering
\caption{Accuracies (\%) for node classification. Best: bold. Runner-up: underline. OOM: Out-of-memory. A.R.: Average rating.}
\label{res_all_acc}
\resizebox{\textwidth}{!}{%
\begin{tabular}{@{}cc|cccc|ccccc@{}}
\toprule
\multicolumn{2}{c|}{Model} & Cora-C & Citeseer & PubMed & Zoo & House & IMDB & Twitch & Amazon & A.R.      \\ \midrule
\multicolumn{1}{c|}{\multirow{10}{*}{\rotatebox{90}{supervised}}}    & GCN       & 73.19$\pm$1.92    & 63.72$\pm$2.20      & 76.32$\pm$1.19       & 42.76$\pm$9.92         & 65.09$\pm$2.29         & 45.02$\pm$2.26     & 49.20$\pm$2.39 & 21.75$\pm$2.19     & 20.0                       \\
\multicolumn{1}{c|}{}                               & GAT       & 73.92$\pm$2.12     & 64.75$\pm$2.52     & 76.15$\pm$0.92          & 42.26$\pm$10.22         & 65.69$\pm$2.42      & 45.39$\pm$2.20         & 49.72$\pm$2.53 & 22.75$\pm$2.52       & 18.9                                  \\
\multicolumn{1}{c|}{}                               & HGNN      & 73.59$\pm$2.92     & 64.69$\pm$2.35     & 79.56$\pm$1.60       & 77.69$\pm$10.29            & 65.66$\pm$2.32     & 49.72$\pm$1.35     & 50.22$\pm$2.59 & 22.15$\pm$2.39       & 16.4                                    \\
\multicolumn{1}{c|}{}                               & HyperGCN  & 74.62$\pm$2.36     & 65.36$\pm$2.59     & 78.16$\pm$1.39       & 78.39$\pm$10.92        & 66.32$\pm$2.82          & 45.45$\pm$2.11       & 50.52$\pm$1.29 & 23.75$\pm$1.19            & 15.0                            \\
\multicolumn{1}{c|}{}                               & HNHN      & 73.66$\pm$2.25     & 65.19$\pm$2.10     & 79.32$\pm$0.96        & 78.89$\pm$10.29          & 67.28$\pm$2.29      & 49.95$\pm$1.72      & 50.12$\pm$2.69 & 23.12$\pm$2.72          & 14.3                             \\
\multicolumn{1}{c|}{}                               & UniGCN    & 75.21$\pm$2.20     & 66.39$\pm$2.16     & 78.19$\pm$0.72           & 78.02$\pm$12.20        & 66.52$\pm$1.92          & 49.20$\pm$1.55      & 50.72$\pm$1.62 & 23.95$\pm$1.39          & 13.9                           \\
\multicolumn{1}{c|}{}                               & AllSet    & 77.15$\pm$1.66     & 67.52$\pm$1.90     & 81.39$\pm$0.62       & 78.19$\pm$10.32       & 67.62$\pm$1.36       & 49.91$\pm$1.22      & 51.02$\pm$2.19 & 23.95$\pm$1.30         & 10.6                                   \\ \cmidrule(l){2-11}
\multicolumn{1}{c|}{}                               & FAGCN &  77.45$\pm$2.32     & 68.33$\pm$2.64    & 82.86$\pm$0.90       & 64.63$\pm$13.45       & 53.66$\pm$4.68       & 56.05$\pm$2.67 & 51.20$\pm$0.64 & 25.66$\pm$0.27 & 10.0 \\
\multicolumn{1}{c|}{}                               & H2GCN &  76.69$\pm$3.19     & 68.92$\pm$2.40    & 83.03$\pm$0.74       & 45.62$\pm$12.24       & 53.57$\pm$4.62       & 56.29$\pm$1.19 & 50.89$\pm$0.65 & 27.03$\pm$2.40 & 10.4 \\
\multicolumn{1}{c|}{}                               & $R^{2}$LP &  45.37$\pm$3.16  & 46.34$\pm$4.88 & 75.50$\pm$1.25    & 63.33$\pm$11.67   & 67.71$\pm$2.64       & 50.59$\pm$1.70 & {\ul 51.56$\pm$0.64} & {\ul 27.23$\pm$0.63} & 13.1\\
\midrule
\multicolumn{1}{c|}{\multirow{12}{*}{\rotatebox{90}{unsupervised}}} & DGI       & 76.22$\pm$1.52    & 66.92$\pm$1.60      & 77.96$\pm$2.29       & 63.19$\pm$13.19         & 66.82$\pm$1.52      & 49.39$\pm$1.76      & 50.71$\pm$1.15 & 24.12$\pm$1.76    & 14.8                \\
\multicolumn{1}{c|}{}                               & GRACE     & 76.99$\pm$2.52    & 66.55$\pm$2.59      & 80.62$\pm$0.69         & 62.29$\pm$15.36          & 66.50$\pm$2.36      & 50.06$\pm$2.12     & 50.79$\pm$1.12 & 24.32$\pm$1.16           & 13.4      \\
\multicolumn{1}{c|}{}                               & BGRL      & 77.12$\pm$1.62    & 67.35$\pm$2.12      & 81.39$\pm$0.99         & 65.55$\pm$12.30     & 66.39$\pm$1.76      & 50.25$\pm$2.19      & 50.62$\pm$1.52 & 24.62$\pm$1.16         & 12.6              \\
\multicolumn{1}{c|}{}                               & COSTA     & 77.59$\pm$1.86    & 68.92$\pm$1.75      & 81.82$\pm$0.56       & 69.76$\pm$10.22            & 67.32$\pm$1.39    & 50.32$\pm$1.69     & 51.17$\pm$1.25 & 24.95$\pm$1.56           & 9.8        \\ \cmidrule(l){2-11} 
\multicolumn{1}{c|}{}                               & GREET  & 79.70$\pm$1.10    & 67.63$\pm$1.78      & {\ul 83.38$\pm$0.52}       & {\ul 80.12$\pm$11.03}         & {\ul73.83$\pm$1.58}    & 56.18$\pm$0.91        
& 51.54$\pm$0.57       & 25.78$\pm$0.33   & {\ul 4.1}          \\
\multicolumn{1}{c|}{}                               & MUSE  & 78.24$\pm$1.14    & 70.10$\pm$1.37     & 80.56$\pm$0.59       & 77.78$\pm$10.43         & 73.03$\pm$1.02   & {\ul 56.42$\pm$1.03}      & 51.50$\pm$0.67     & OOM       & 7.3   \\ \cmidrule(l){2-11}
\multicolumn{1}{c|}{}                               & HyperGRL  & 75.02$\pm$1.39    &  69.10$\pm$1.62     & 82.29$\pm$0.72      & 70.19$\pm$11.32    & 68.95$\pm$2.09      & 50.76$\pm$0.90       & 51.21$\pm$0.65 & 25.22$\pm$0.76           & 9.4              \\
\multicolumn{1}{c|}{}                               & HyperGCL  & 73.01$\pm$1.55    & 69.72$\pm$1.55      & 82.95$\pm$0.38       & 64.69$\pm$12.68         & 70.26$\pm$2.25    & 53.19$\pm$0.75     & 51.37$\pm$0.62 & 25.22$\pm$0.56         & 9.5                \\
\multicolumn{1}{c|}{}                               & TriCL     & 80.36$\pm$1.17    & {\ul 72.06$\pm$1.19}      & 83.26$\pm$0.52   & 79.37$\pm$11.17  & 72.28$\pm$1.54    & 52.51$\pm$0.93  & 51.42$\pm$0.57 & 26.22$\pm$0.32       & 4.5         \\
\multicolumn{1}{c|}{}                               & HypeBoy   & 75.29$\pm$1.77    & 70.59$\pm$1.79      & 75.87$\pm$0.94               & 75.92$\pm$11.46        &  72.66$\pm$1.97          & 52.91$\pm$1.45       & 51.39$\pm$0.70 & 26.35$\pm$0.39   & 8.8                   \\
\multicolumn{1}{c|}{}                               & MMACL     & {\ul 80.81$\pm$1.00}  &  71.59$\pm$1.44  & 82.03$\pm$0.64        & 74.81$\pm$8.86      & 72.69$\pm$1.13     & 53.92$\pm$0.83    & OOM & OOM          & 9.4     \\ 
\cmidrule(l){2-11} 

\multicolumn{1}{c|}{}                           &  HONOR    & \textbf{81.42$\pm$1.15} & \textbf{73.23$\pm$1.00}  & \textbf{83.69$\pm$0.66} & \textbf{82.78$\pm$7.13} & \textbf{75.73$\pm$1.10} & \textbf{58.37$\pm$0.86} 
& \textbf{51.82$\pm$0.27}  & \textbf{30.62$\pm$0.35} & \textbf{1.0}\\ \bottomrule
\end{tabular}%
}
\end{table*}

\begin{table*}[ht]
\centering
\caption{NMI and ARI of different models for node clustering tasks.} 
\tiny
\label{res_nmi_ari}
\resizebox{\textwidth}{!}{%
\begin{tabular}{@{}c|cccccccc|ccccccccc@{}}
\toprule
\multirow{2}{*}{Model} & \multicolumn{2}{c}{Cora-C}   & \multicolumn{2}{c}{Citeseer}   & \multicolumn{2}{c}{PubMed}      & \multicolumn{2}{c|}{Zoo}      & \multicolumn{2}{c}{House}      & \multicolumn{2}{c}{IMDB}      & \multicolumn{2}{c}{Twitch}       & \multicolumn{2}{c}{Amazon}       & \multirow{2}{*}{A.R.}    \\ \cmidrule(l){2-17} 
                       & NMI            & ARI            & NMI            & ARI            & NMI            & ARI            & NMI            & ARI            & NMI            & ARI            & NMI            & ARI            & NMI           & ARI            & NMI             & ARI                   \\ \midrule
DGI                    &  {\ul 52.82}    &  {\ul 46.60} & 31.62 & 23.70   &  30.42          & 26.32           & 15.09          & 10.26     & 10.56          & 10.21    & 3.79          & 3.92                              & 0.06 & 0.05 & 0.32 & 0.15 & 8.3  \\
GRACE                  & 44.96          & 32.29      & 32.90 & 25.29    & 16.75          & 16.55              & 12.39          & 10.16     & 11.25          & 10.72  
& 3.86          & 3.99                     & 0.09 & 0.07 & 0.29 & 0.12   & 9.5 \\
BGRL                   & 41.29          & 22.59      & 33.19 & 26.29    & 16.22          & 17.29            & 72.55          & 60.65       & 11.55          & 10.59      & 4.20          & 4.36                         & 0.10 & 0.02 & 0.30 & 0.19 & 9.0 \\
COSTA                  & 47.76          & 36.39     & 35.19 & 28.22     & 20.59          & 16.02              & 76.25          & 61.95     & 12.29          & 10.40     & 4.12          & 4.20                             & 0.13 & 0.06  & 0.35 & 0.16 & 7.9 \\ \midrule

GREET                  & 46.13          & 32.79     & 41.53 & 41.17     & 16.89          & 12.26              & {\ul 92.20}         & {\ul 87.62}     & 9.90          & 8.53     & 5.61          & 3.88 & 0.10 & 0.01 & 0.32 & 0.15 & 7.9\\

MUSE                  & 48.24         & 37.35     & 41.70 & 41.91     & 20.19          & 18.00              & 84.86          & 85.83    & 13.81          & {\ul 13.32}     & 5.73         & 5.01 & 0.00 & -0.01 & OOM &  OOM & 6.2\\

\midrule
HyperGRL               & 41.79          & 35.09      & 37.19  & 35.12    & 27.90          & 23.35         & 78.95          & 79.62    & 10.79          & 10.36     & 5.19          & 4.76  & 0.13 & 0.09 & 0.69 & 0.45       & 6.9    \\
HyperGCL               & 49.33          & 39.64       & 39.12 & 38.15   & 28.49          &  27.73         & 71.49          & 66.00   & 13.07          & 10.42      &  6.19             &  8.17 & 0.16 & 0.10 & 0.72 & {\ul 0.55}                                    &  4.9\\
TriCL                  & 52.40          & 44.43    &  {\ul 42.60} & 41.52      & 25.61          & 20.62         &  89.50    &  85.25    & 12.85          & 10.78    & 5.85          & 5.93                     &  {\ul 0.17} & 0.01 &  0.63 & 0.32  & {\ul 4.6}\\
HypeBoy               & 42.68          & 29.03    & 38.25 & 36.39      & {\ul 31.02}   & {\ul 28.06}             & 79.29          & 82.22         &  13.21              &  12.29                &  8.11             & {\ul 9.32}             & 0.15 & {\ul 0.11} & {\ul 0.79} & 0.39         &  {\ul 4.6} \\
MMACL                  & 52.68          & 42.25     & 42.48 &  \textbf{42.46}      & 24.34          & 18.81            & 80.79          & 81.11    &  {\ul 15.00}    & 11.27   &  {\ul 8.39}          &  8.41       & OOM & OOM  & OOM & OOM                &  5.5    \\ \midrule
HONOR                   
              & \textbf{53.62} & \textbf{51.81}   
              & \textbf{42.68} & {\ul 42.03} 
              & \textbf{35.23} & \textbf{39.18} 
              & \textbf{94.48} & \textbf{90.33} 
              & \textbf{20.55} & \textbf{27.17} 
              & \textbf{11.42} & \textbf{12.84} 
              & \textbf{0.23} & \textbf{0.19} 
              & \textbf{2.31} & \textbf{0.69}  
              & \textbf{1.1} \\ \bottomrule
\end{tabular}%
}
\end{table*}

\noindent \textbf{Model Performance.}
According to the results in Tables \ref{res_all_acc} and \ref{res_nmi_ari}, HONOR consistently demonstrates highly competitive performance compared to all kinds of baselines. 
This is mainly attributed to the incorporation of a high-pass filtering mechanism in HONOR’s design, which effectively enhances the decoupling of heterophilic structures between hyperedges and nodes, thereby avoiding the oversmoothing issue that traditional low-pass filters may cause in heterophilic hypergraphs. Meanwhile, by integrating a dual-view contrastive learning strategy, our model can extract complementary information from different structural relationships.

We also observe that compared to other types of models, heterophilic graph contrastive learning models generally perform better on the evaluated datasets. This is mainly because these methods are specifically designed to learn discriminative node representations for heterophilic graphs. However, their performance remains inferior to ours, which may be attributed to their complete neglect of high-order interactions in hypergraphs. As for hypergraph contrastive learning models, their training paradigm enables them to learn distinctive node representations. However, these methods rely on homophily assumption in their design, and thus their overall performance does not reach the level of heterophilic graph contrastive learning models. 

\begin{table*}[ht]
\centering
\caption{Accuracies (\%) of different model variants on the datasets. Best: bold. A.R.: Average rating.}
\label{ablation}
\tiny
\resizebox{0.9\textwidth}{!}{%
\begin{tabular}{@{}c|cccc|ccccc@{}}
\toprule
Method  & Cora-C & Citeseer & PubMed & Zoo & House & IMDB & Twitch & Amazon & A.R.  \\
\midrule
\textit{w/o Dec.}   & 75.27$\pm$0.77 & 71.16$\pm$1.06 & 82.32$\pm$0.83 & 77.65$\pm$7.41 & 73.69$\pm$8.87 & 56.46$\pm$1.25  & 51.53$\pm$0.61 & 28.77$\pm$0.37 & 6.8 \\

\textit{w/o Cov.}  & 81.02$\pm$1.15 & 73.20$\pm$0.99 & 83.68$\pm$0.64 & 82.65$\pm$7.26 & 75.46$\pm$2.05 & 58.27$\pm$0.83  & 51.60$\pm$0.59 & 26.32$\pm$0.64 & 3.6 \\

\textit{w/o Pro.} & 78.78$\pm$1.19 & 68.37$\pm$1.49 & 83.16$\pm$0.91 & 81.48$\pm$7.22 & 72.60$\pm$3.62 & 53.64$\pm$1.02 & 51.51$\pm$0.55 & 29.97$\pm$0.24 & 6.4 \\

\textit{w/o Deg.} & \textbf{81.56$\pm$1.13} & 72.66$\pm$1.01 & 83.48$\pm$0.73 & 79.69$\pm$10.73 & 70.01$\pm$3.66 & 58.35$\pm$1.03 & 50.65$\pm$0.62 & 30.57$\pm$0.34 & 4.6\\

\midrule

\textit{Mean-based} & 81.38$\pm$1.12 & 73.21$\pm$1.01 & 83.68$\pm$0.64 & 82.59$\pm$6.93 & 74.06$\pm$3.42 & 58.31$\pm$0.87 & 51.25$\pm$0.69 & 30.31$\pm$0.24 & 3.5\\

\textit{Low-pass} & 79.13$\pm$1.37 & 67.75$\pm$2.19 & 83.25$\pm$0.54 & 78.83$\pm$11.21 & 50.56$\pm$4.96 & 55.08$\pm$0.71 & 50.71$\pm$0.59 & 27.71$\pm$0.54 & 7.5\\

\midrule

\textit{Pro. Rep.}  & 81.32$\pm$1.12 & 72.71$\pm$1.11 & 83.65$\pm$0.64 & 80.49$\pm$10.75 & 74.54$\pm$2.06 & 56.59$\pm$0.91  & 51.35$\pm$0.60 & 30.41$\pm$0.25 & 4.1\\

\textit{Att. Rep.} & 80.19$\pm$1.20 & 71.24$\pm$1.65 & 82.92$\pm$0.35 & 77.78$\pm$10.88 & 75.33$\pm$0.80 & 55.77$\pm$0.95  & 51.21$\pm$0.57 & 27.46$\pm$0.69 & 6.5\\

\midrule

HONOR & 81.42$\pm$1.15 & \textbf{73.23$\pm$1.00}  & \textbf{83.69$\pm$0.66} & \textbf{82.78$\pm$7.13} & \textbf{75.73$\pm$1.10} & \textbf{58.37$\pm$0.86} 
& \textbf{51.82$\pm$0.27}  & \textbf{30.62$\pm$0.35} & \textbf{1.1}
\\
\bottomrule
\end{tabular}
}
\end{table*}

\noindent \textbf{Ablation Study.}
To assess the necessity of the employed techniques, we design a series of model variants applied for all the evaluated datasets. The results are shown in Table \ref{ablation}. \textit{First}, we separately remove the following key components: the decoupling module (\textit{w/o Dec.}), the covariance regularization term (\textit{w/o Cov.}), the prompt-based mechanism (\textit{w/o Pro.}), and the degree-based node structural augmentation (\textit{w/o Deg.}). The results show that each module contributes to the overall performance, with the decoupling module and the prompt-based mechanism making the most significant contributions. \textit{Subsequently}, we replace the two key designs (attention-based view generation mechanism and high-pass filter) with alternative schemes (mean-based view generation mechanism and GCN-based low-pass filter). And the results indicates that replacing any of the key designs will lead to a performance decrease. Notably, the results of the low-pass variant show that when the high-pass filter is replaced with a low-pass filter (GCN), the average ranking (7.5) is the worst among all variants, which further highlights the critical role of the high-pass filter in capturing local heterophily. \textit{Finally}, we further investigate the quality of representations generated by the prompt-based (\textit{Pro. Rep.}) and attention-based (\textit{Att. Rep.}) view generation mechanisms. It can be observed that representations generated from a single view are inferior to those produced by the final dual-view combination. 
\begin{figure*}[ht]
    \centering
    \begin{subfigure}[b]{0.25\textwidth}
        \centering
        \raisebox{0.35cm}{\includegraphics[width=\textwidth]{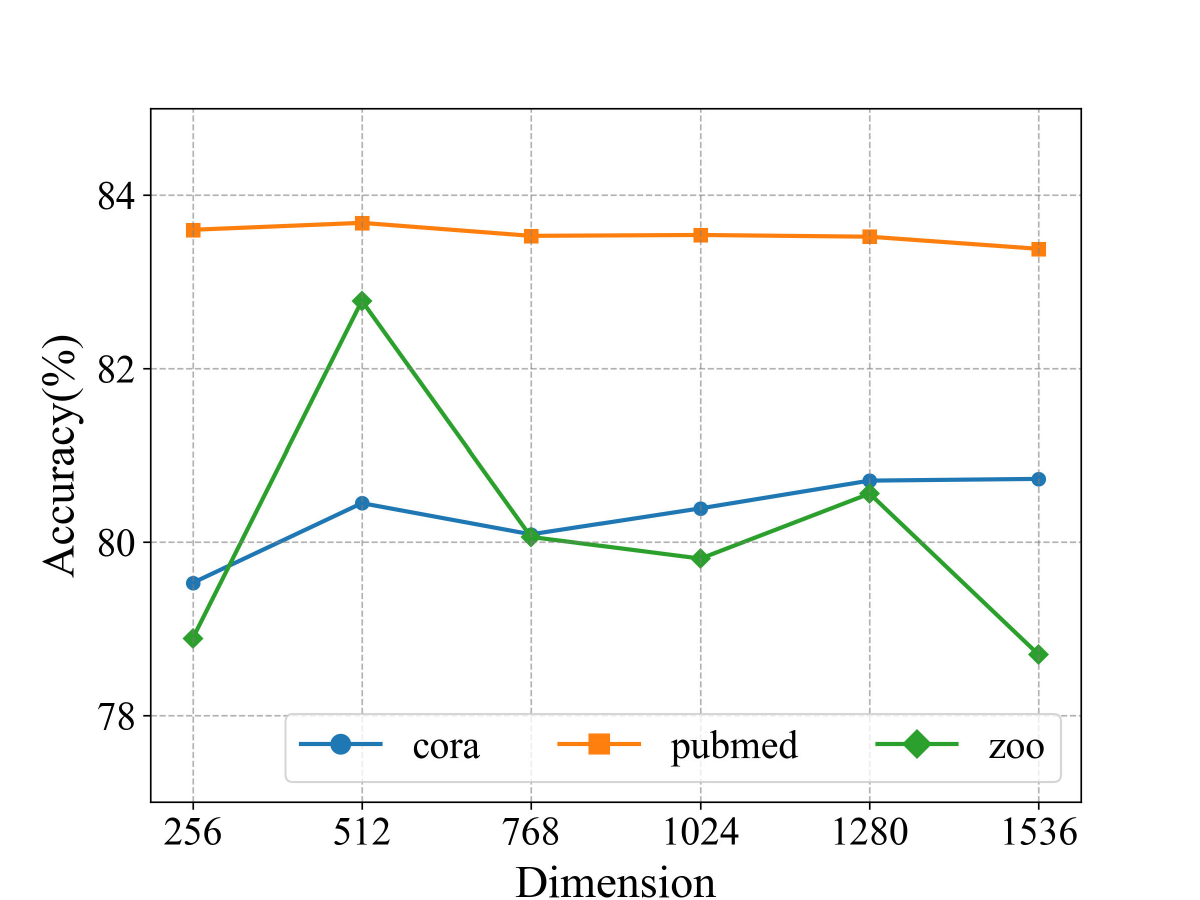}}
        \caption{$D$ (homophilic datasets)}
        \label{sub1}
    \end{subfigure}
    \begin{subfigure}[b]{0.25\textwidth}
        \centering
        \raisebox{0.35cm}
        {\includegraphics[width=\textwidth]{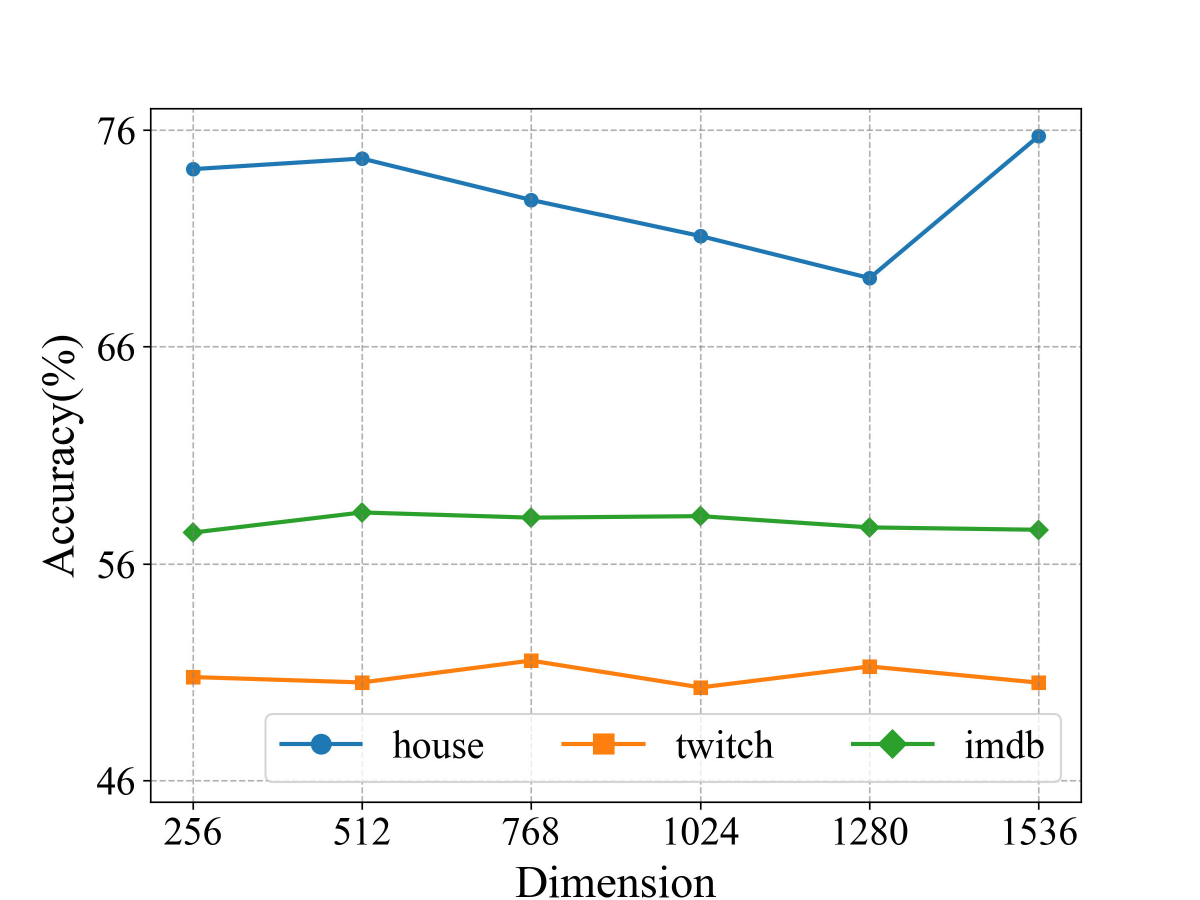}}
        \caption{$D$ (heterophilic datasets)}
        \label{sub2}
    \end{subfigure}
    \begin{subfigure}[b]{0.21\textwidth}
        \centering
        \includegraphics[width=\textwidth]{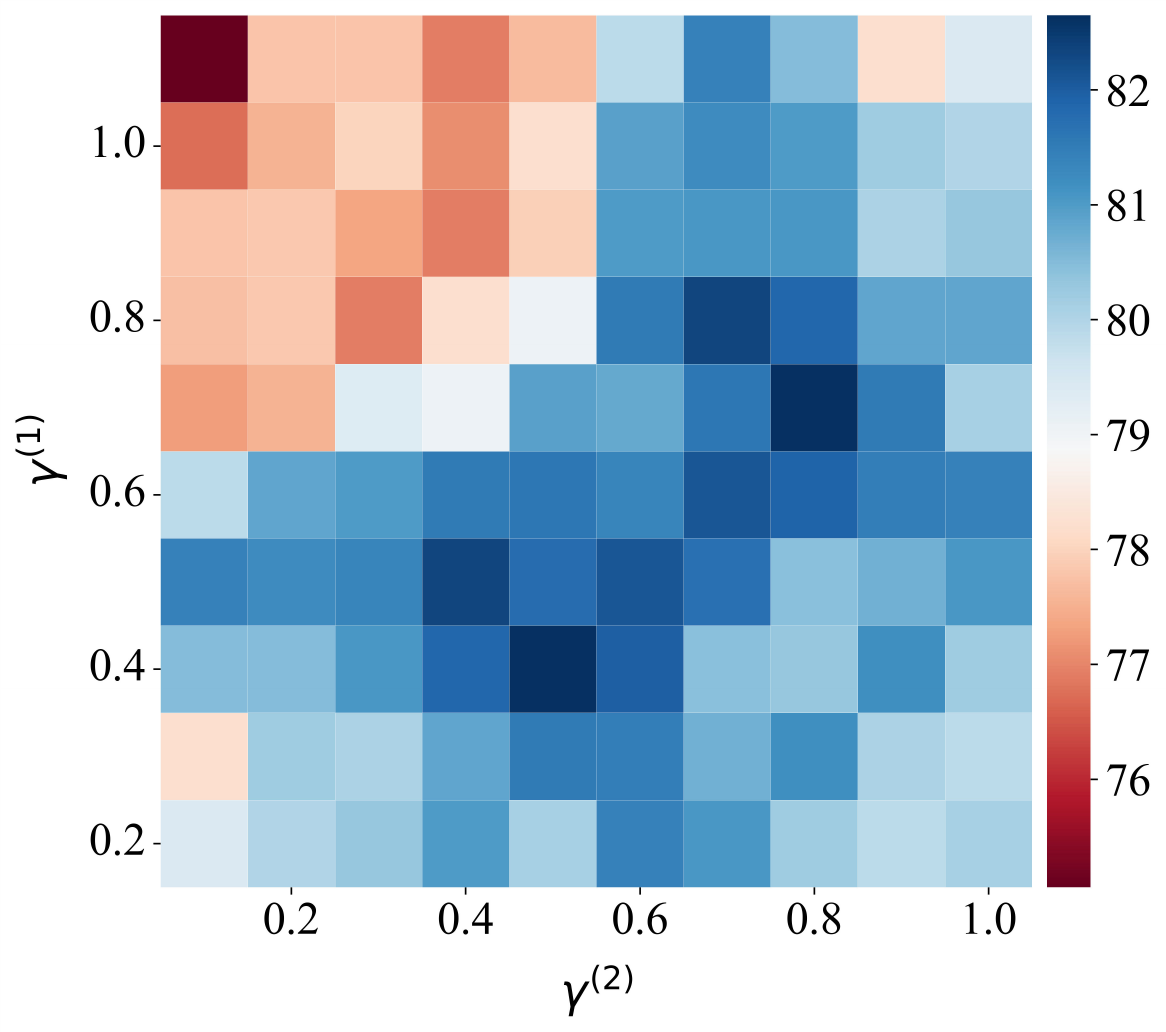}
        \caption{$\gamma^{(1)}$ and $\gamma^{(2)}$ (Zoo)}
        \label{sub3}
    \end{subfigure}
    \begin{subfigure}[b]{0.21\textwidth}
        \centering
        \includegraphics[width=\textwidth]{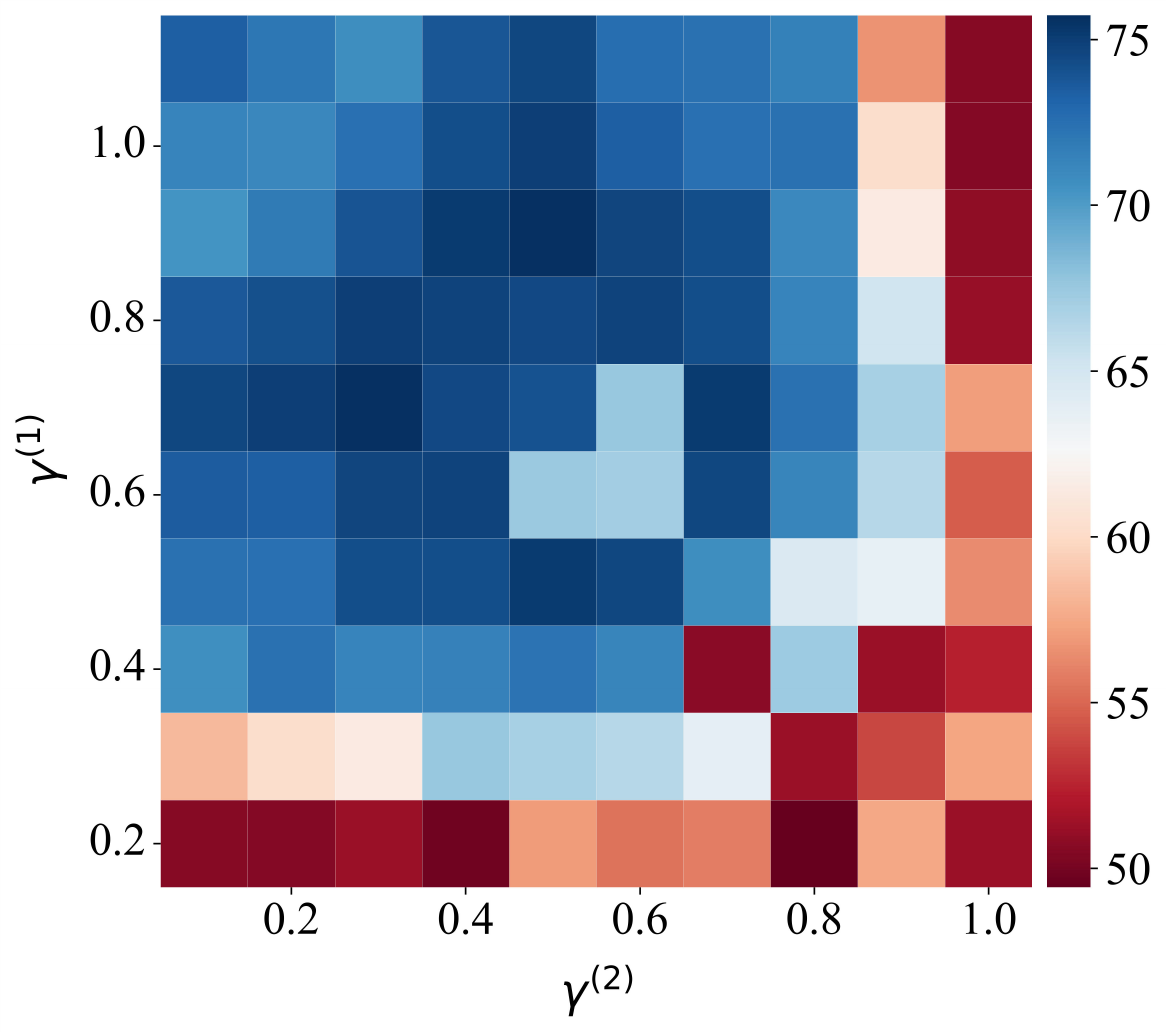}
        \caption{$\gamma^{(1)}$ and $\gamma^{(2)}$ (House)}
        \label{sub4}
    \end{subfigure}
    \caption{Hyperparameter sensitivity for node classification tasks across multiple datasets.} 
    \label{hyperparameter}
\end{figure*}

\begin{table*}[!ht]
\centering
\caption{Statistical analysis of mean, variance, and entropy before and after training.}
\label{high-pass role}
\resizebox{0.9\textwidth}{!}{%
\begin{tabular}{@{}c|cccccccccccccccccc@{}}
\toprule
\multirow{2}{*}{Dataset}             & \multicolumn{3}{c}{House}        & \multicolumn{3}{c}{IMDB}       & \multicolumn{3}{c}{Twitch}         & \multicolumn{3}{c}{Amazon}     & \multicolumn{3}{c}{PubMed}      & \multicolumn{3}{c}{Zoo}   \\ \cmidrule(l){2-19} 
                       & Mean$\downarrow$   & Var$\uparrow$ & Entropy$\uparrow$    & Mean$\downarrow$ & Var$\uparrow$   & Entropy$\uparrow$ & Mean$\downarrow$   & Var$\uparrow$ & Entropy$\uparrow$   & Mean$\downarrow$ & Var$\uparrow$   & Entropy$\uparrow$ & Mean$\downarrow$   & Var$\uparrow$ & Entropy$\uparrow$   & Mean$\downarrow$ & Var$\uparrow$   & Entropy$\uparrow$ \\ \midrule
Before         & 0.998          & 0          & 0.1         & 0.992          & 0          & 0.179          & 0.979          & 0.003          & 1.157          & 0.931          & 0.002          & 2.956         & 0.998          & 0          & 0.003             & 0.976            & 0.001          & 1.615                    \\
After          & \textbf{0.66}            & \textbf{0.237}            & \textbf{4.524}          & \textbf{0.722}          & \textbf{0.019}          & \textbf{4.78}          & \textbf{0.903}          & \textbf{0.007}          & \textbf{3.609}          & \textbf{0.81}          & \textbf{0.02}          & \textbf{4.46}            & \textbf{0.897}         & \textbf{0.017}         & \textbf{3.736}             & \textbf{0.478}            & \textbf{0.214}          & \textbf{5.246}                \\ \bottomrule
\end{tabular}%
}
\end{table*}

\noindent \textbf{Hyperparameter Sensitivity.}
We investigate two key hyperparameters that affect the proposed HONOR's performance: the hidden dimension $D$ of the high-pass encoder, and the high-pass filtering coefficient $\gamma^{(1)}$ and $\gamma^{(2)}$. 
\\
\textit{Hidden dimension $D$.}
We first analyze the impact of the hidden dimension on model performance using several homophilic and heterophilic datasets, respectively. In this experiment, we keep other hyperparameters fixed, and the results are shown in Figs. \ref{hyperparameter} (a) and (b). We find that for homophilic datasets, as the hidden dimension increases, the model performance generally improves. 
However, further increasing the hidden dimension do not lead to a continuous improvement. 
For heterophilic datasets, simply increasing the hidden dimension may even introduce noise or overfitting. Nevertheless, even in this scenario, our model maintains stable performance across different dimensions, demonstrating its robustness to heterophilic structures.
\\
\textit{High-pass filtering coefficient $\gamma^{(1)}$ and $\gamma^{(2)}$.}
We fix other hyperparameters and adjust $\gamma^{(1)}$ and $\gamma^{(2)}$ from 0.1 to 1.0 with a step size of 0.1. Experiments are conducted on the Zoo and House datasets, and the results are shown in Figs. \ref{hyperparameter} (c) and (d). The results indicate that the optimal combination of high-pass filtering coefficients varies significantly depending on the data structure: for Zoo, better performance is achieved when $\gamma^{(1)}$ is relatively small and $\gamma^{(2)}$ is moderately large; in contrast, for House, a larger $\gamma^{(1)}$ and a smaller $\gamma^{(2)}$ lead to better performance. This phenomenon demonstrates that we can adjust the high-pass filtering coefficients according to the homophily or heterophily of the hypergraph, controlling the degree of high-frequency information extraction.

\noindent \textbf{Effectiveness of High-Pass Filtering.}
To further assess the effectiveness of the high-pass filter in enhancing the decoupling of heterophilic structures within hyperedge-node relationships, we additionally measure the cosine similarity distribution between node embeddings and their connected hyperedges, and conduct a comparative analysis \textbf{before} and \textbf{after} training. Specifically, we report three key metrics: the mean similarity (Mean), the variance of similarity (Var) and the average local entropy (Entropy). The mean similarity is defined as 
\( \mu = \frac{1}{N} \sum_{i=1}^{N} s_i \),
where \( s_i \) denotes the cosine similarity of the \( i \)-th hyperedge-node pair and \( N \) is the total number of pairs.
The variance is computed as 
\( \nu^2 = \frac{1}{N} \sum_{i=1}^{N} (s_i - \mu)^2 \),
capturing the spread of similarity values.
The average local entropy is given by 
\( H = -\frac{1}{N} \sum_{i=1}^{N} \sum_{j} p_{ij} \log p_{ij} \), where \( p_{ij} \) is the normalized similarity (obtained via softmax) between node \( i \) and its connected hyperedges. 
As shown in Table \ref{high-pass role}, the results demonstrate that after introducing the high-pass filter, the model’s mean similarity decreases significantly, while the similarity variance and local entropy increase overall. This confirms the effectiveness of the high-pass filter in preserving structural heterophily, enhancing representational capacity, and preventing representation collapse.

\section{Conclusion}

In this paper, we propose HONOR, an unsupervised contrastive learning framework designed for both homophilic and heterophilic hypergraphs, aiming to overcome the limitations of existing methods that rely on the homophily assumption. We first introduce a novel heterophily discrimination metric, named Label Entropy, to assess the degree of heterophily in real-world hypergraphs quantitatively. Then, HONOR explicitly models the heterophilic relationships between hyperedges and their connected nodes through two complementary hyperedge feature construction mechanisms: prompt-based and attention-based aggregation. 
Supported by high-pass filtering, this design enables HONOR to fully exploit heterophilic connection patterns in hypergraphs, producing more discriminative representations. 
We validate the effectiveness of HONOR from both theoretical analysis and empirical evaluations. More importantly, this study provides new insights for handling complex and diverse hypergraph structures and offers valuable inspiration for the design of future hypergraph learning frameworks.

\bibliographystyle{ACM-Reference-Format}
\balance
\bibliography{sample-base}


\begin{thebibliography}{50}


\ifx \showCODEN    \undefined \def \showCODEN     #1{\unskip}     \fi
\ifx \showISBNx    \undefined \def \showISBNx     #1{\unskip}     \fi
\ifx \showISBNxiii \undefined \def \showISBNxiii  #1{\unskip}     \fi
\ifx \showISSN     \undefined \def \showISSN      #1{\unskip}     \fi
\ifx \showLCCN     \undefined \def \showLCCN      #1{\unskip}     \fi
\ifx \shownote     \undefined \def \shownote      #1{#1}          \fi
\ifx \showarticletitle \undefined \def \showarticletitle #1{#1}   \fi
\ifx \showURL      \undefined \def \showURL       {\relax}        \fi
\providecommand\bibfield[2]{#2}
\providecommand\bibinfo[2]{#2}
\providecommand\natexlab[1]{#1}
\providecommand\showeprint[2][]{arXiv:#2}

\bibitem[Abu-El-Haija et~al\mbox{.}(2019)]%
        {abu2019mixhop}
\bibfield{author}{\bibinfo{person}{Sami Abu-El-Haija}, \bibinfo{person}{Bryan Perozzi}, \bibinfo{person}{Amol Kapoor}, \bibinfo{person}{Nazanin Alipourfard}, \bibinfo{person}{Kristina Lerman}, \bibinfo{person}{Hrayr Harutyunyan}, \bibinfo{person}{Greg Ver~Steeg}, {and} \bibinfo{person}{Aram Galstyan}.} \bibinfo{year}{2019}\natexlab{}.
\newblock \showarticletitle{Mixhop: Higher-order graph convolutional architectures via sparsified neighborhood mixing}. In \bibinfo{booktitle}{\emph{ICML}}.
\newblock


\bibitem[Antelmi et~al\mbox{.}(2023)]%
        {antelmi2023survey}
\bibfield{author}{\bibinfo{person}{Alessia Antelmi}, \bibinfo{person}{Gennaro Cordasco}, \bibinfo{person}{Mirko Polato}, \bibinfo{person}{Vittorio Scarano}, \bibinfo{person}{Carmine Spagnuolo}, {and} \bibinfo{person}{Dingqi Yang}.} \bibinfo{year}{2023}\natexlab{}.
\newblock \showarticletitle{A survey on hypergraph representation learning}.
\newblock \bibinfo{journal}{\emph{Comput. Surveys}} \bibinfo{volume}{56}, \bibinfo{number}{1} (\bibinfo{year}{2023}), \bibinfo{pages}{1--38}.
\newblock


\bibitem[Asuncion et~al\mbox{.}(2007)]%
        {asuncion2007uci}
\bibfield{author}{\bibinfo{person}{Arthur Asuncion}, \bibinfo{person}{David Newman}, {et~al\mbox{.}}} \bibinfo{year}{2007}\natexlab{}.
\newblock \bibinfo{title}{UCI machine learning repository}.
\newblock


\bibitem[Bo et~al\mbox{.}(2021)]%
        {bo2021beyond}
\bibfield{author}{\bibinfo{person}{Deyu Bo}, \bibinfo{person}{Xiao Wang}, \bibinfo{person}{Chuan Shi}, {and} \bibinfo{person}{Huawei Shen}.} \bibinfo{year}{2021}\natexlab{}.
\newblock \showarticletitle{Beyond low-frequency information in graph convolutional networks}. In \bibinfo{booktitle}{\emph{AAAI}}.
\newblock


\bibitem[Cheng et~al\mbox{.}(2024)]%
        {cheng2024resurrecting}
\bibfield{author}{\bibinfo{person}{Yao Cheng}, \bibinfo{person}{Caihua Shan}, \bibinfo{person}{Yifei Shen}, \bibinfo{person}{Xiang Li}, \bibinfo{person}{Siqiang Luo}, {and} \bibinfo{person}{Dongsheng Li}.} \bibinfo{year}{2024}\natexlab{}.
\newblock \showarticletitle{Resurrecting label propagation for graphs with heterophily and label noise}. In \bibinfo{booktitle}{\emph{SIGKDD}}.
\newblock


\bibitem[Chien et~al\mbox{.}(2022)]%
        {chien2021you}
\bibfield{author}{\bibinfo{person}{Eli Chien}, \bibinfo{person}{Chao Pan}, \bibinfo{person}{Jianhao Peng}, {and} \bibinfo{person}{Olgica Milenkovic}.} \bibinfo{year}{2022}\natexlab{}.
\newblock \showarticletitle{You are allset: A multiset function framework for hypergraph neural networks}. In \bibinfo{booktitle}{\emph{ICLR}}.
\newblock


\bibitem[Chien et~al\mbox{.}(2021)]%
        {chien2020adaptive}
\bibfield{author}{\bibinfo{person}{Eli Chien}, \bibinfo{person}{Jianhao Peng}, \bibinfo{person}{Pan Li}, {and} \bibinfo{person}{Olgica Milenkovic}.} \bibinfo{year}{2021}\natexlab{}.
\newblock \showarticletitle{Adaptive universal generalized pagerank graph neural network}.
\newblock \bibinfo{journal}{\emph{ICLR}} (\bibinfo{year}{2021}).
\newblock


\bibitem[Dong et~al\mbox{.}(2020)]%
        {dong2020hnhn}
\bibfield{author}{\bibinfo{person}{Yihe Dong}, \bibinfo{person}{Will Sawin}, {and} \bibinfo{person}{Yoshua Bengio}.} \bibinfo{year}{2020}\natexlab{}.
\newblock \showarticletitle{Hnhn: Hypergraph networks with hyperedge neurons}.
\newblock \bibinfo{journal}{\emph{ICML}} (\bibinfo{year}{2020}).
\newblock


\bibitem[Du et~al\mbox{.}(2022)]%
        {du2022self}
\bibfield{author}{\bibinfo{person}{Boxin Du}, \bibinfo{person}{Changhe Yuan}, \bibinfo{person}{Robert Barton}, \bibinfo{person}{Tal Neiman}, {and} \bibinfo{person}{Hanghang Tong}.} \bibinfo{year}{2022}\natexlab{}.
\newblock \showarticletitle{Self-supervised hypergraph representation learning}. In \bibinfo{booktitle}{\emph{IEEE Big Data}}.
\newblock


\bibitem[Feng et~al\mbox{.}(2019)]%
        {feng2019hypergraph}
\bibfield{author}{\bibinfo{person}{Yifan Feng}, \bibinfo{person}{Haoxuan You}, \bibinfo{person}{Zizhao Zhang}, \bibinfo{person}{Rongrong Ji}, {and} \bibinfo{person}{Yue Gao}.} \bibinfo{year}{2019}\natexlab{}.
\newblock \showarticletitle{Hypergraph neural networks}. In \bibinfo{booktitle}{\emph{AAAI}}.
\newblock


\bibitem[Glorot and Bengio(2010)]%
        {glorot2010understanding}
\bibfield{author}{\bibinfo{person}{Xavier Glorot} {and} \bibinfo{person}{Yoshua Bengio}.} \bibinfo{year}{2010}\natexlab{}.
\newblock \showarticletitle{Understanding the difficulty of training deep feedforward neural networks}. In \bibinfo{booktitle}{\emph{AISTATS}}.
\newblock


\bibitem[He et~al\mbox{.}(2023)]%
        {he2023contrastive}
\bibfield{author}{\bibinfo{person}{Dongxiao He}, \bibinfo{person}{Jitao Zhao}, \bibinfo{person}{Rui Guo}, \bibinfo{person}{Zhiyong Feng}, \bibinfo{person}{Di Jin}, \bibinfo{person}{Yuxiao Huang}, \bibinfo{person}{Zhen Wang}, {and} \bibinfo{person}{Weixiong Zhang}.} \bibinfo{year}{2023}\natexlab{}.
\newblock \showarticletitle{Contrastive learning meets homophily: two birds with one stone}. In \bibinfo{booktitle}{\emph{ICML}}.
\newblock


\bibitem[Huang and Yang(2021)]%
        {huang2021unignn}
\bibfield{author}{\bibinfo{person}{Jing Huang} {and} \bibinfo{person}{Jie Yang}.} \bibinfo{year}{2021}\natexlab{}.
\newblock \showarticletitle{Unignn: a unified framework for graph and hypergraph neural networks}. In \bibinfo{booktitle}{\emph{IJCAI}}.
\newblock


\bibitem[Jiang et~al\mbox{.}(2022)]%
        {jiang2022hypergraph}
\bibfield{author}{\bibinfo{person}{Ping Jiang}, \bibinfo{person}{Xiaoheng Deng}, \bibinfo{person}{Leilei Wang}, \bibinfo{person}{Zailiang Chen}, {and} \bibinfo{person}{Shichao Zhang}.} \bibinfo{year}{2022}\natexlab{}.
\newblock \showarticletitle{Hypergraph representation for detecting 3D objects from noisy point clouds}.
\newblock \bibinfo{journal}{\emph{IEEE Transactions on Knowledge and Data Engineering}} \bibinfo{volume}{35}, \bibinfo{number}{7} (\bibinfo{year}{2022}), \bibinfo{pages}{7016--7029}.
\newblock


\bibitem[Kim et~al\mbox{.}(2018)]%
        {kim2018stochastic}
\bibfield{author}{\bibinfo{person}{Chiheon Kim}, \bibinfo{person}{Afonso~S Bandeira}, {and} \bibinfo{person}{Michel~X Goemans}.} \bibinfo{year}{2018}\natexlab{}.
\newblock \showarticletitle{Stochastic block model for hypergraphs: Statistical limits and a semidefinite programming approach}.
\newblock \bibinfo{journal}{\emph{arXiv preprint arXiv:1807.02884}} (\bibinfo{year}{2018}).
\newblock


\bibitem[Kim et~al\mbox{.}(2022)]%
        {kim2022equivariant}
\bibfield{author}{\bibinfo{person}{Jinwoo Kim}, \bibinfo{person}{Saeyoon Oh}, \bibinfo{person}{Sungjun Cho}, {and} \bibinfo{person}{Seunghoon Hong}.} \bibinfo{year}{2022}\natexlab{}.
\newblock \showarticletitle{Equivariant hypergraph neural networks}. In \bibinfo{booktitle}{\emph{ECCV}}.
\newblock


\bibitem[Kim et~al\mbox{.}(2024a)]%
        {kim2024hypeboy}
\bibfield{author}{\bibinfo{person}{Sunwoo Kim}, \bibinfo{person}{Shinhwan Kang}, \bibinfo{person}{Fanchen Bu}, \bibinfo{person}{Soo~Yong Lee}, \bibinfo{person}{Jaemin Yoo}, {and} \bibinfo{person}{Kijung Shin}.} \bibinfo{year}{2024}\natexlab{a}.
\newblock \showarticletitle{Hypeboy: Generative self-supervised representation learning on hypergraphs}. In \bibinfo{booktitle}{\emph{ICLR}}.
\newblock


\bibitem[Kim et~al\mbox{.}(2024b)]%
        {kim2024survey}
\bibfield{author}{\bibinfo{person}{Sunwoo Kim}, \bibinfo{person}{Soo~Yong Lee}, \bibinfo{person}{Yue Gao}, \bibinfo{person}{Alessia Antelmi}, \bibinfo{person}{Mirko Polato}, {and} \bibinfo{person}{Kijung Shin}.} \bibinfo{year}{2024}\natexlab{b}.
\newblock \showarticletitle{A survey on hypergraph neural networks: an in-depth and step-by-step guide}. In \bibinfo{booktitle}{\emph{SIGKDD}}.
\newblock


\bibitem[Kingma and Ba(2014)]%
        {kingma2014adam}
\bibfield{author}{\bibinfo{person}{Diederik~P. Kingma} {and} \bibinfo{person}{Jimmy Ba}.} \bibinfo{year}{2014}\natexlab{}.
\newblock \showarticletitle{Adam: A method for stochastic optimization}. In \bibinfo{booktitle}{\emph{ICLR}}.
\newblock


\bibitem[Kipf and Welling(2017)]%
        {kipf2016semi}
\bibfield{author}{\bibinfo{person}{Thomas~N Kipf} {and} \bibinfo{person}{Max Welling}.} \bibinfo{year}{2017}\natexlab{}.
\newblock \showarticletitle{Semi-supervised classification with graph convolutional networks}. In \bibinfo{booktitle}{\emph{ICLR}}.
\newblock


\bibitem[Lee and Shin(2023)]%
        {lee2023m}
\bibfield{author}{\bibinfo{person}{Dongjin Lee} {and} \bibinfo{person}{Kijung Shin}.} \bibinfo{year}{2023}\natexlab{}.
\newblock \showarticletitle{I’m me, we’re us, and i’m us: Tri-directional contrastive learning on hypergraphs}. In \bibinfo{booktitle}{\emph{AAAI}}.
\newblock


\bibitem[Lee et~al\mbox{.}(2024)]%
        {lee2024survey}
\bibfield{author}{\bibinfo{person}{Geon Lee}, \bibinfo{person}{Fanchen Bu}, \bibinfo{person}{Tina Eliassi-Rad}, {and} \bibinfo{person}{Kijung Shin}.} \bibinfo{year}{2024}\natexlab{}.
\newblock \showarticletitle{A survey on hypergraph mining: Patterns, tools, and generators}. In \bibinfo{booktitle}{\emph{SIGKDD}}.
\newblock


\bibitem[Lee and Chae(2024)]%
        {lee2024multi}
\bibfield{author}{\bibinfo{person}{Jongsoo Lee} {and} \bibinfo{person}{Dong-Kyu Chae}.} \bibinfo{year}{2024}\natexlab{}.
\newblock \showarticletitle{Multi-view Mixed Attention for Contrastive Learning on Hypergraphs}. In \bibinfo{booktitle}{\emph{SIGIR}}.
\newblock


\bibitem[Li et~al\mbox{.}(2024)]%
        {li2024hypergraph}
\bibfield{author}{\bibinfo{person}{Fan Li}, \bibinfo{person}{Xiaoyang Wang}, \bibinfo{person}{Dawei Cheng}, \bibinfo{person}{Wenjie Zhang}, \bibinfo{person}{Ying Zhang}, {and} \bibinfo{person}{Xuemin Lin}.} \bibinfo{year}{2024}\natexlab{}.
\newblock \showarticletitle{Hypergraph Self-supervised Learning with Sampling-efficient Signals}. In \bibinfo{booktitle}{\emph{IJCAI}}.
\newblock


\bibitem[Li et~al\mbox{.}(2025)]%
        {li2025hypergraph}
\bibfield{author}{\bibinfo{person}{Ming Li}, \bibinfo{person}{Yongchun Gu}, \bibinfo{person}{Yi Wang}, \bibinfo{person}{Yujie Fang}, \bibinfo{person}{Lu Bai}, \bibinfo{person}{Xiaosheng Zhuang}, {and} \bibinfo{person}{Pietro Lio}.} \bibinfo{year}{2025}\natexlab{}.
\newblock \showarticletitle{When hypergraph meets heterophily: New benchmark datasets and baseline}. In \bibinfo{booktitle}{\emph{AAAI}}. \bibinfo{pages}{18377--18384}.
\newblock


\bibitem[Liu et~al\mbox{.}(2023a)]%
        {liu2023pre}
\bibfield{author}{\bibinfo{person}{Pengfei Liu}, \bibinfo{person}{Weizhe Yuan}, \bibinfo{person}{Jinlan Fu}, \bibinfo{person}{Zhengbao Jiang}, \bibinfo{person}{Hiroaki Hayashi}, {and} \bibinfo{person}{Graham Neubig}.} \bibinfo{year}{2023}\natexlab{a}.
\newblock \showarticletitle{Pre-train, prompt, and predict: A systematic survey of prompting methods in natural language processing}.
\newblock \bibinfo{journal}{\emph{ACM CSUR}} \bibinfo{volume}{55}, \bibinfo{number}{9} (\bibinfo{year}{2023}), \bibinfo{pages}{1--35}.
\newblock


\bibitem[Liu et~al\mbox{.}(2023b)]%
        {liu2023beyond}
\bibfield{author}{\bibinfo{person}{Yixin Liu}, \bibinfo{person}{Yizhen Zheng}, \bibinfo{person}{Daokun Zhang}, \bibinfo{person}{Vincent~CS Lee}, {and} \bibinfo{person}{Shirui Pan}.} \bibinfo{year}{2023}\natexlab{b}.
\newblock \showarticletitle{Beyond smoothing: Unsupervised graph representation learning with edge heterophily discriminating}. In \bibinfo{booktitle}{\emph{AAAI}}.
\newblock


\bibitem[Ma et~al\mbox{.}(2022)]%
        {ma2021homophily}
\bibfield{author}{\bibinfo{person}{Yao Ma}, \bibinfo{person}{Xiaorui Liu}, \bibinfo{person}{Neil Shah}, {and} \bibinfo{person}{Jiliang Tang}.} \bibinfo{year}{2022}\natexlab{}.
\newblock \showarticletitle{Is homophily a necessity for graph neural networks?}. In \bibinfo{booktitle}{\emph{ICLR}}.
\newblock


\bibitem[Pei et~al\mbox{.}({[n.\,d.]})]%
        {peigeom}
\bibfield{author}{\bibinfo{person}{Hongbin Pei}, \bibinfo{person}{Bingzhe Wei}, \bibinfo{person}{Kevin Chen-Chuan Chang}, \bibinfo{person}{Yu Lei}, {and} \bibinfo{person}{Bo Yang}.} \bibinfo{year}{[n.\,d.]}\natexlab{}.
\newblock \showarticletitle{Geom-GCN: Geometric Graph Convolutional Networks}. In \bibinfo{booktitle}{\emph{ICLR}}.
\newblock


\bibitem[Sen et~al\mbox{.}(2008)]%
        {sen2008collective}
\bibfield{author}{\bibinfo{person}{Prithviraj Sen}, \bibinfo{person}{Galileo Namata}, \bibinfo{person}{Mustafa Bilgic}, \bibinfo{person}{Lise Getoor}, \bibinfo{person}{Brian Galligher}, {and} \bibinfo{person}{Tina Eliassi-Rad}.} \bibinfo{year}{2008}\natexlab{}.
\newblock \showarticletitle{Collective classification in network data}.
\newblock \bibinfo{journal}{\emph{AI Magazine}} \bibinfo{volume}{29}, \bibinfo{number}{3} (\bibinfo{year}{2008}), \bibinfo{pages}{93--93}.
\newblock


\bibitem[Sun et~al\mbox{.}(2023)]%
        {su2023graph}
\bibfield{author}{\bibinfo{person}{Xiangguo Sun}, \bibinfo{person}{Jiawen Zhang}, \bibinfo{person}{Xixi Wu}, \bibinfo{person}{Hong Cheng}, \bibinfo{person}{Yun Xiong}, {and} \bibinfo{person}{Jia Li}.} \bibinfo{year}{2023}\natexlab{}.
\newblock \showarticletitle{Graph prompt learning: A comprehensive survey and beyond}.
\newblock \bibinfo{journal}{\emph{arXiv preprint arXiv:2311.16534}} (\bibinfo{year}{2023}).
\newblock


\bibitem[Thakoor et~al\mbox{.}(2021)]%
        {thakoor2021bootstrapped}
\bibfield{author}{\bibinfo{person}{Shantanu Thakoor}, \bibinfo{person}{Corentin Tallec}, \bibinfo{person}{Mohammad~Gheshlaghi Azar}, \bibinfo{person}{R{\'e}mi Munos}, \bibinfo{person}{Petar Veli{\v{c}}kovi{\'c}}, {and} \bibinfo{person}{Michal Valko}.} \bibinfo{year}{2021}\natexlab{}.
\newblock \showarticletitle{Bootstrapped representation learning on graphs}. In \bibinfo{booktitle}{\emph{ICLR}}.
\newblock


\bibitem[Velickovic et~al\mbox{.}(2018)]%
        {velickovic2017graph}
\bibfield{author}{\bibinfo{person}{Petar Velickovic}, \bibinfo{person}{Guillem Cucurull}, \bibinfo{person}{Arantxa Casanova}, \bibinfo{person}{Adriana Romero}, \bibinfo{person}{Pietro Lio}, \bibinfo{person}{Yoshua Bengio}, {et~al\mbox{.}}} \bibinfo{year}{2018}\natexlab{}.
\newblock \showarticletitle{Graph attention networks}. In \bibinfo{booktitle}{\emph{ICLR}}.
\newblock


\bibitem[Veli{\v{c}}kovi{\'c} et~al\mbox{.}(2019)]%
        {velivckovic2018deep}
\bibfield{author}{\bibinfo{person}{Petar Veli{\v{c}}kovi{\'c}}, \bibinfo{person}{William Fedus}, \bibinfo{person}{William~L Hamilton}, \bibinfo{person}{Pietro Li{\`o}}, \bibinfo{person}{Yoshua Bengio}, {and} \bibinfo{person}{R~Devon Hjelm}.} \bibinfo{year}{2019}\natexlab{}.
\newblock \showarticletitle{Deep graph infomax}. In \bibinfo{booktitle}{\emph{ICLR}}.
\newblock


\bibitem[Wang et~al\mbox{.}(2023)]%
        {wang2022equivariant}
\bibfield{author}{\bibinfo{person}{Peihao Wang}, \bibinfo{person}{Shenghao Yang}, \bibinfo{person}{Yunyu Liu}, \bibinfo{person}{Zhangyang Wang}, {and} \bibinfo{person}{Pan Li}.} \bibinfo{year}{2023}\natexlab{}.
\newblock \showarticletitle{Equivariant hypergraph diffusion neural operators}. In \bibinfo{booktitle}{\emph{ICLR}}.
\newblock


\bibitem[Wang et~al\mbox{.}(2019)]%
        {wang2019heterogeneous}
\bibfield{author}{\bibinfo{person}{Xiao Wang}, \bibinfo{person}{Houye Ji}, \bibinfo{person}{Chuan Shi}, \bibinfo{person}{Bai Wang}, \bibinfo{person}{Yanfang Ye}, \bibinfo{person}{Peng Cui}, {and} \bibinfo{person}{Philip~S Yu}.} \bibinfo{year}{2019}\natexlab{}.
\newblock \showarticletitle{Heterogeneous graph attention network}. In \bibinfo{booktitle}{\emph{WWW}}.
\newblock


\bibitem[Wang et~al\mbox{.}(2021)]%
        {wang2021multi}
\bibfield{author}{\bibinfo{person}{Yingheng Wang}, \bibinfo{person}{Yaosen Min}, \bibinfo{person}{Xin Chen}, {and} \bibinfo{person}{Ji Wu}.} \bibinfo{year}{2021}\natexlab{}.
\newblock \showarticletitle{Multi-view graph contrastive representation learning for drug-drug interaction prediction}. In \bibinfo{booktitle}{\emph{WWW}}.
\newblock


\bibitem[Wang et~al\mbox{.}(2022)]%
        {wang2022acmp}
\bibfield{author}{\bibinfo{person}{Yuelin Wang}, \bibinfo{person}{Kai Yi}, \bibinfo{person}{Xinliang Liu}, \bibinfo{person}{Yu~Guang Wang}, {and} \bibinfo{person}{Shi Jin}.} \bibinfo{year}{2022}\natexlab{}.
\newblock \showarticletitle{ACMP: Allen-cahn message passing with attractive and repulsive forces for graph neural networks}. In \bibinfo{booktitle}{\emph{ICLR}}.
\newblock


\bibitem[Wei et~al\mbox{.}(2022)]%
        {wei2022augmentations}
\bibfield{author}{\bibinfo{person}{Tianxin Wei}, \bibinfo{person}{Yuning You}, \bibinfo{person}{Tianlong Chen}, \bibinfo{person}{Yang Shen}, \bibinfo{person}{Jingrui He}, {and} \bibinfo{person}{Zhangyang Wang}.} \bibinfo{year}{2022}\natexlab{}.
\newblock \showarticletitle{Augmentations in hypergraph contrastive learning: Fabricated and generative}. In \bibinfo{booktitle}{\emph{NeurIPS}}.
\newblock


\bibitem[Wu et~al\mbox{.}(2024)]%
        {wu2024collaborative}
\bibfield{author}{\bibinfo{person}{Hanrui Wu}, \bibinfo{person}{Nuosi Li}, \bibinfo{person}{Jia Zhang}, \bibinfo{person}{Sentao Chen}, \bibinfo{person}{Michael~K Ng}, {and} \bibinfo{person}{Jinyi Long}.} \bibinfo{year}{2024}\natexlab{}.
\newblock \showarticletitle{Collaborative contrastive learning for hypergraph node classification}.
\newblock \bibinfo{journal}{\emph{PR}} (\bibinfo{year}{2024}).
\newblock


\bibitem[Xia et~al\mbox{.}(2022)]%
        {xia2022self}
\bibfield{author}{\bibinfo{person}{Lianghao Xia}, \bibinfo{person}{Chao Huang}, {and} \bibinfo{person}{Chuxu Zhang}.} \bibinfo{year}{2022}\natexlab{}.
\newblock \showarticletitle{Self-supervised hypergraph transformer for recommender systems}. In \bibinfo{booktitle}{\emph{SIGKDD}}.
\newblock


\bibitem[Xia et~al\mbox{.}(2021)]%
        {xia2021self}
\bibfield{author}{\bibinfo{person}{Xin Xia}, \bibinfo{person}{Hongzhi Yin}, \bibinfo{person}{Junliang Yu}, \bibinfo{person}{Qinyong Wang}, \bibinfo{person}{Lizhen Cui}, {and} \bibinfo{person}{Xiangliang Zhang}.} \bibinfo{year}{2021}\natexlab{}.
\newblock \showarticletitle{Self-supervised hypergraph convolutional networks for session-based recommendation}. In \bibinfo{booktitle}{\emph{AAAI}}.
\newblock


\bibitem[Xiao et~al\mbox{.}(2022)]%
        {xiao2022decoupled}
\bibfield{author}{\bibinfo{person}{Teng Xiao}, \bibinfo{person}{Zhengyu Chen}, \bibinfo{person}{Zhimeng Guo}, \bibinfo{person}{Zeyang Zhuang}, {and} \bibinfo{person}{Suhang Wang}.} \bibinfo{year}{2022}\natexlab{}.
\newblock \showarticletitle{Decoupled self-supervised learning for non-homophilous graphs}.
\newblock \bibinfo{journal}{\emph{NeurIPS}} (\bibinfo{year}{2022}).
\newblock


\bibitem[Yadati et~al\mbox{.}(2019)]%
        {yadati2019hypergcn}
\bibfield{author}{\bibinfo{person}{Naganand Yadati}, \bibinfo{person}{Madhav Nimishakavi}, \bibinfo{person}{Prateek Yadav}, \bibinfo{person}{Vikram Nitin}, \bibinfo{person}{Anand Louis}, {and} \bibinfo{person}{Partha Talukdar}.} \bibinfo{year}{2019}\natexlab{}.
\newblock \showarticletitle{Hypergcn: A new method for training graph convolutional networks on hypergraphs}. In \bibinfo{booktitle}{\emph{NeurIPS}}.
\newblock


\bibitem[Yuan et~al\mbox{.}(2023)]%
        {yuan2023muse}
\bibfield{author}{\bibinfo{person}{Mengyi Yuan}, \bibinfo{person}{Minjie Chen}, {and} \bibinfo{person}{Xiang Li}.} \bibinfo{year}{2023}\natexlab{}.
\newblock \showarticletitle{Muse: multi-view contrastive learning for heterophilic graphs}. In \bibinfo{booktitle}{\emph{ACM}}.
\newblock


\bibitem[Zhang et~al\mbox{.}(2022)]%
        {zhang2022costa}
\bibfield{author}{\bibinfo{person}{Yifei Zhang}, \bibinfo{person}{Hao Zhu}, \bibinfo{person}{Zixing Song}, \bibinfo{person}{Piotr Koniusz}, {and} \bibinfo{person}{Irwin King}.} \bibinfo{year}{2022}\natexlab{}.
\newblock \showarticletitle{COSTA: covariance-preserving feature augmentation for graph contrastive learning}. In \bibinfo{booktitle}{\emph{SIGKDD}}.
\newblock


\bibitem[Zheng et~al\mbox{.}(2019)]%
        {zheng2019gene}
\bibfield{author}{\bibinfo{person}{Xiao Zheng}, \bibinfo{person}{Wenyang Zhu}, \bibinfo{person}{Chang Tang}, {and} \bibinfo{person}{Minhui Wang}.} \bibinfo{year}{2019}\natexlab{}.
\newblock \showarticletitle{Gene selection for microarray data classification via adaptive hypergraph embedded dictionary learning}.
\newblock \bibinfo{journal}{\emph{Gene}}  \bibinfo{volume}{706} (\bibinfo{year}{2019}), \bibinfo{pages}{188--200}.
\newblock


\bibitem[Zhu et~al\mbox{.}(2023)]%
        {zhu2023heterophily}
\bibfield{author}{\bibinfo{person}{Jiong Zhu}, \bibinfo{person}{Yujun Yan}, \bibinfo{person}{Mark Heimann}, \bibinfo{person}{Lingxiao Zhao}, \bibinfo{person}{Leman Akoglu}, {and} \bibinfo{person}{Danai Koutra}.} \bibinfo{year}{2023}\natexlab{}.
\newblock \showarticletitle{Heterophily and graph neural networks: Past, present and future}.
\newblock \bibinfo{journal}{\emph{IEEE Data Engineering Bulletin}} (\bibinfo{year}{2023}).
\newblock


\bibitem[Zhu et~al\mbox{.}(2020b)]%
        {zhu2020beyond}
\bibfield{author}{\bibinfo{person}{Jiong Zhu}, \bibinfo{person}{Yujun Yan}, \bibinfo{person}{Lingxiao Zhao}, \bibinfo{person}{Mark Heimann}, \bibinfo{person}{Leman Akoglu}, {and} \bibinfo{person}{Danai Koutra}.} \bibinfo{year}{2020}\natexlab{b}.
\newblock \showarticletitle{Beyond homophily in graph neural networks: Current limitations and effective designs}.
\newblock \bibinfo{journal}{\emph{Advances in neural information processing systems}}  \bibinfo{volume}{33} (\bibinfo{year}{2020}), \bibinfo{pages}{7793--7804}.
\newblock


\bibitem[Zhu et~al\mbox{.}(2020a)]%
        {zhu2020deep}
\bibfield{author}{\bibinfo{person}{Yanqiao Zhu}, \bibinfo{person}{Yichen Xu}, \bibinfo{person}{Feng Yu}, \bibinfo{person}{Qiang Liu}, \bibinfo{person}{Shu Wu}, {and} \bibinfo{person}{Liang Wang}.} \bibinfo{year}{2020}\natexlab{a}.
\newblock \showarticletitle{Deep graph contrastive representation learning}.
\newblock \bibinfo{journal}{\emph{ICML}} (\bibinfo{year}{2020}).
\newblock


\end{thebibliography}


\end{document}